\documentclass{article}

\usepackage[preprint]{neurips_2025}

\usepackage[utf8]{inputenc} % allow utf-8 input
\usepackage[T1]{fontenc}    % use 8-bit T1 fonts
\usepackage{hyperref}       % hyperlinks
\usepackage{url}            % simple URL typesetting
\usepackage{booktabs}       % professional-quality tables
\usepackage{amsfonts}       % blackboard math symbols
\usepackage{nicefrac}       % compact symbols for 1/2, etc.
\usepackage{microtype}      % microtypography
\usepackage{xcolor}         % colors

\usepackage{kotex}
\usepackage{setspace}

\usepackage{amsmath}
\usepackage{amssymb}
\usepackage{mathtools}
\usepackage{amsthm}
\usepackage[capitalise]{cleveref}
\usepackage[ruled,vlined]{algorithm2e}
\usepackage{thmtools,thm-restate}
\usepackage{cancel}
\usepackage{pifont}

\usepackage{algorithmic}
\usepackage{multicol,multirow}
\usepackage{subcaption}
\usepackage{adjustbox}
\usepackage{float}
\usepackage{svg}

\usepackage{thm-restate}
\newtheorem{theorem}{Theorem}[section]

\newtheorem{lemma}[theorem]{Lemma}

\newtheorem{definition}[theorem]{Definition}

\newtheorem{remark}[theorem]{Remark}

\crefname{algocfline}{Algorithm}{Algorithms}

\title{Catalyst: a Novel Regularizer for Structured Pruning with Auxiliary Extension of Parameter Space
}

\author{%
  Jaeheun Jung,\\
  Department of Mathematics\\
  Korea University\\
  Seoul, Republic of Korea \\
  \texttt{wodsos@korea.ac.kr} \\
  \And
  Donghun Lee\thanks{corresponding author}\\
  Department of Mathematics\\
  Korea University\\
  Seoul, Republic of Korea \\
  \texttt{holy@korea.ac.kr} \\
}

\begin{document}

\maketitle

\begin{abstract}
Structured pruning aims to reduce the size and computational cost of deep neural networks by removing entire filters or channels.
The traditional regularizers such as L1 or Group Lasso and its variants lead to magnitude-biased pruning decisions, such that the filters with small magnitudes are likely to be pruned. Also, they often entail pruning results with almost zero margin around pruning decision boundary, such that tiny perturbation in a filter magnitude can flip the pruning decision.
In this paper, we identify the precise algebraic condition under which pruning operations preserve model performance, and use the condition to construct a novel regularizer defined in an extended parameter space via auxiliary catalyst variables.
The proposed Catalyst regularization ensures fair pruning chance for each filters with theoretically provable zero bias to their magnitude and robust pruning behavior achieved by wide-margin bifurcation of magnitudes between the preserved and the pruned filters.
The theoretical properties naturally lead to real-world effectiveness, as shown by empirical validations of Catalyst Pruning algorithm. Pruning results on various datasets and models are superior to state-of-the-art filter pruning methods, and at the same time confirm the predicted robust and fair pruning characteristics of Catalyst pruning.
\end{abstract}

\section{Introduction}

Structured pruning \cite{he2023structured} is a widely used technique to reduce the computational cost of deep neural networks by removing redundant filters. 
Regularization-based approaches have gained popularity for their simplicity and compatibility with standard model training pipelines. 
Classical regularization methods such as $L_1$ or Group Lasso \cite{yuan2006grouplasso,wen2016ssl} promote sparse pruning by penalizing the magnitude of model parameters. These pruning techniques, however, exhibit a critical limitation, as they are more likely to prune filters with smaller norms regardless of their true contribution to the model performance. 
The pruning bias in magnitude is particularly problematic in pretrained models where filters with small weights often affects the model outcome significantly \cite{He_2019_CVPR}. 

The magnitude bias in pruning has deeper roots: conventional regularizers have geometric misalignment to the pruning-invariant set, a set of parameters where pruning does not change the model output. 
Pruning with misaligned magnitude-based regularizations have risk of pushing filter indiscriminately toward the origin and compromising important filters. 
Recent approaches restrict the parameter space to identify filters for pruning early and apply regularization selectively \cite{jiang2023pruning,wang2019structured}, thereby avoiding excessive sparsification of essential components and avoiding the degenerate global minima of misaligned regularization. 
However, these improvements still do not address the root cause: the inherent magnitude-driven pruning bias. 

Moreover, traditional regularizers fail to express a robust decision boundary between pruned and preserved filters. 
In such settings, small perturbations in filter norms can lead to unstable or inconsistent pruning decisions. 
Some methods attempt to address this by encouraging prune-or-preserve bifurcation in proxy variables such as filter norms, gates, or masks \cite{zhuang2020polarization,guo2021gdp,ma2022differentiable}, leading filters to cluster around distinct ``preserve'' or ``prune'' decisions. 
However, these approaches often fail to elicit robust bifurcation, as their heuristics tend to make sensitive pruning decisions with narrow gap between pruned and preserved filters. 

To address these challenges, we formally characterize the algebraic conditions for lossless structured pruning, under which pruning preserves the neural network model. 
Using this foundation, we construct a novel regularizer in an extended parameter space by introducing auxiliary diagonal ``catalyst'' variables. 
We propose Catalyst, a novel algebraically grounded regularizer for structured pruning, and show that it has many desirable properties:
\begin{itemize}
    \item Fair pruning due to provably zero magnitude bias: Catalyst provides fair pruning opportunity across filters by decoupling pruning decisions from the magnitude bias.
    \item Robust pruning by wide-margin bifurcation: Catalyst naturally induces a wide margin between prune and preserve decisions, increasing stability and interpretability of pruning.
    \item Empirical effectiveness backed by theoretical rigor: Catalyst demonstrates consistently superior pruning performance across multiple benchmarks, as predicted by its perfect geometric alignment to pruning-invariant set.
\end{itemize}

\section{Related Works}\label{sec:relworks}

\subsection{Structured pruning}
Structured pruning methods remove one or more filters, which is slice of target weight parameter along the axis of output dimension, from the given model.
Structured pruning methods usually exploit $L_1$ or Group Lasso \cite{yuan2006grouplasso,wen2016ssl} regularizers to select filters to prune from diverse target layers such as batch normalization layers \cite{liu2017slim,zhuang2020polarization,kang2020scp}, convolution layers \cite{wang2020greg,wu2024ato}, external layers \cite{ding2021resrep,guo2021gdp} or multiple filters from different layers \cite{fang2023depgraph}. 
Our work defines the algebraic condition to achieve lossless pruning and proposes a novel algebraically principled regularizer for structured pruning methods.

To make pruning decisions more robust, structured pruning methods often accompany additional strategies to encourage bifurcation of the values that serve as proxies for pruning decision. 
\cite{zhuang2020polarization} introduces neuron polarization regularizer to bifurcate the filter norms, and more recent works introduce polarization on gates \cite{guo2021gdp} or pruning masks \cite{mo2023pruning,ma2022differentiable}.
Unlike prior works that introduce explicit strategies to encourage bifurcation, our work enjoys provably robust bifurcation based on the algebraically principled regularizer itself, and demonstrates robust bifurcation behaviors befitting the theoretical analysis.

\subsection{Bypassing algorithm as a structure reparameterization}

The training process consist of reparametrization-train-contraction chain have been proposed\cite{ding2021repvgg,hu22OREPA,zhang2023repnas,fei2024repan,ding2021resrep} and evaluated in literature of deep learning for specific purposes.

Recently, new type of structural reparameterization was proposed in bypassing \cite{bypass} which aim to escape the stationary points driven by SGD, by following 3 stages: 1) extending the model by modifying activation, 
2) training with algebraic constraint, called comeback constraint, and 
3) contracting back to the original model during the training. The comeback constraint of bypassing\cite{bypass} is designed to ensure the lossless contraction.

Unlike conventional structural reparameterization methods, 1) their extension exploits minimal increase in parameter counts and extends the representative ability of the model in function space and 2) the contraction would be lossless only if the comeback constraint is satisfied, while the Rep family can always transform their model to deploy.

\section{Theoretical Aspects}\label{sec:sparsity_reg}

We first formulate the structured pruning problem in \cref{subsec:problem_formulation}. 
Then, in \cref{subsec:algebraic} we define the condition for lossless pruning and construct a novel algebraically principled regularizer. 
In \cref{subsec:bypassing}, Bypass pipeline is adapted for structured pruning.
In \cref{subsec:bifurcation}, we present initialization technique for the proposed method with a theoretical analysis of the robust bifurcation dynamics of the novel regularizer. % and the initialization method on additional parameters.

\subsection{Problem Formulation: Structured Pruning }\label{subsec:problem_formulation}

We consider the problem of pruning a neural network model $\varphi_1:\theta\mapsto f_\theta$
which maps parameters $\theta$ in domain $\mathcal{D}_1$ to a specific function $f_\theta$ in function space $\mathcal{F}$, and includes the subnetwork $\overline{\varphi_1}$ defined below.
\newline
\begin{definition}
We define $\overline{\varphi_1}$ as a deep neural network model that contains two layers, which are considered as a submodule. 
For such a submodule, we use the following notation:
\begin{equation}
    \overline{\varphi_1}(W,b_W,A,b_A) = NN(W,b_W,A,b_A,\sigma)
\end{equation}
where
\begin{equation}
    \begin{aligned}
        NN(W,b_W,A,b_A,\sigma):\mathbb{R}^{N_I}&\rightarrow\mathbb{R}^{N_A}\\
            x&\mapsto b_A+A\sigma(b_W+Wx)
    \end{aligned}
\end{equation}
such that
\begin{enumerate}
    \item $N_I$ and $N_A$ are the dimension of input vector and output vector of $\overline{\varphi_1}$, respectively. We also denote the dimension of hidden vector $Wx+b_W$ by $N_W$.
    \item $W,b_W$ are the weight and bias parameters of the target layer. The $Wx$ may refer to the linear functional parametrized by $W$ such as convolution $Conv(W,x)$ or batch normalization with scaling factor $W$. The addition of $b_W$ may be broadcast on all channels.
    \item $A,b_A$ are the weight and bias parameters of the next layer, defined similarly to $W$ and $b_W$. 
    \item $\sigma$ is the element-wise (channel-wise) operation, which may consist of multiple layers such as activation functions, depthwise convolution, pooling and normalizations. 
\end{enumerate}
\end{definition}

\begin{definition}\label{def:structured pruning}
    For given neural network $NN(W,b_W,A,b_A,\sigma)$ and target layer $W$, we define followings.
    \begin{enumerate}
        \item We define filter $F_i=W_{i,:}$ for $i\in [N_w]$, the parameter which determines the $i$th channel of hidden feature.
        \item Let $P\in 2^{N_W}$ be the set of filter indices. We denote $W[P]$ be the tensor obtained by replacing each $F_i$ to 0 if $i\notin P$. 
        Note that the map $W\mapsto W[P^c]$ becomes out-channel pruning operation on $W$, when $P$ is the set of pruning indices. The in-channel pruning operation would be denoted by $A\mapsto (A^T[P^c])^T$.
        \item Filter Pruning is to minimize 
        \begin{equation}
            \mathcal{L}(\varphi_1(\tilde{\theta},W[P^c],b_W[P^c],(A^T[P^c])^T,b_A))
        \end{equation} with $|P|\geq 1$, where $\tilde{\theta}$ represents the parameters of $\varphi_1$ which are independent to $\overline{\varphi_1}$.
    \end{enumerate}
\end{definition}

\subsection{Equation of lossless pruning}\label{subsec:algebraic}

\begin{figure*}[t]
  \centering
  \begin{subfigure}[b]{0.3\textwidth}
    \includegraphics[width=\textwidth]{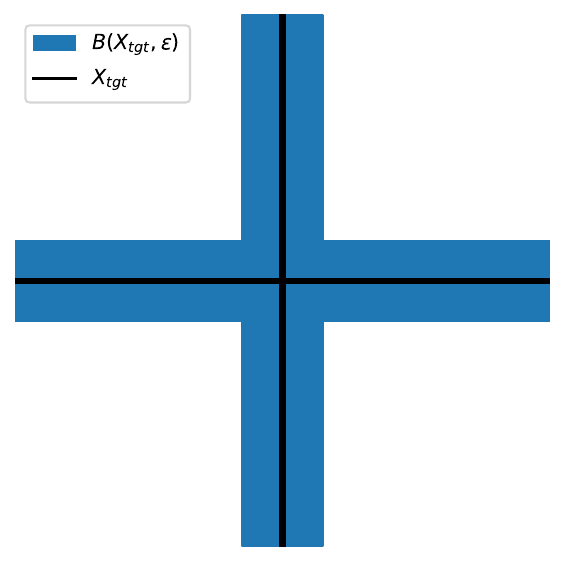}
    \caption{$X_{tgt}$ and $\epsilon$-neighborhood}
    \label{subfig:intuitive_target}
    \end{subfigure}
    \begin{subfigure}[b]{0.3\textwidth}
      \includegraphics[width=\textwidth]{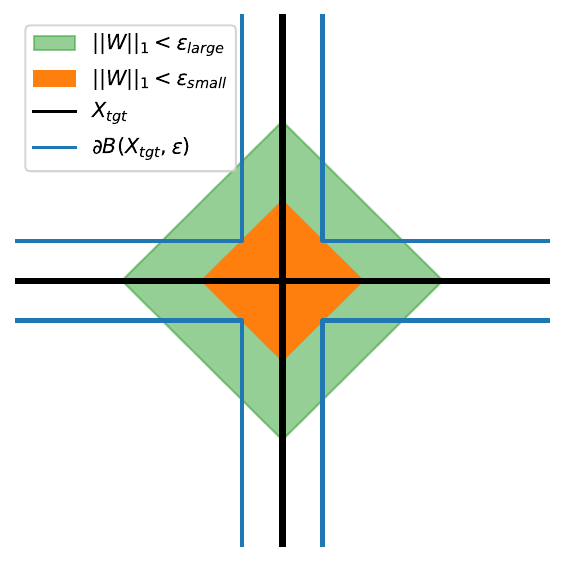}
      \caption{$\epsilon$-closure of $L_1$}
      \label{subfig:intuitive_L1}
    \end{subfigure}
    \begin{subfigure}[b]{0.3\textwidth}
      \includegraphics[width=\textwidth]{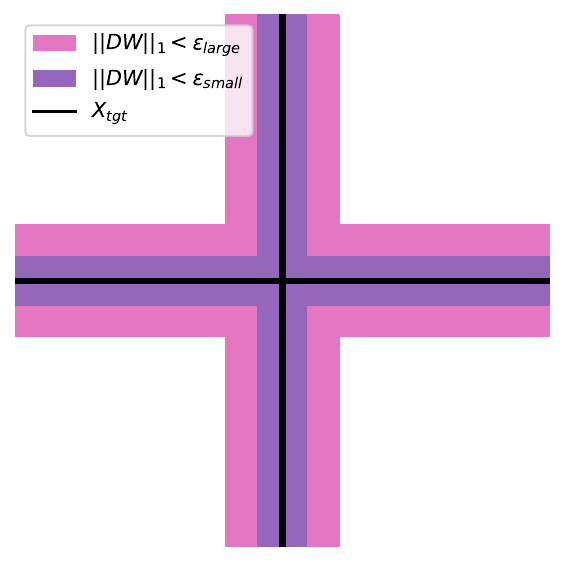}
      \caption{$\epsilon$-closure of Ours}
      \label{subfig:intuitive_DW}
    \end{subfigure}

\caption{Intuitive plot of geometric mismatch in structured pruning. If $W$ has two channels where individuals are 1-dimensional, the pruning-invariant set $X_{tgt}$ would be each axis (black-colored crossing lines).}
\label{fig:intuitive}
\end{figure*}

Structured pruning eliminates selected channels by zeroing out the corresponding filters. While existing regularization methods attempt to minimize the damage due to pruning by sparsifying the filter parameters, they fall short in one crucial aspect: the vicinity of the global minima of these regularizers does not geometrically align with what we term the pruning-invariant set. We define this pruning-invariant set, $X^{tgt}$ , as a finite union of linear subspaces.
\begin{equation}
    X_{tgt}:=\bigcup_{i = 1}^{N_W}\{W|W_{i,:}=F_i=0\}.
\end{equation}
 If $W$ is in $X_{tgt}$, then $i$th filter $F_i$ is zero for some $i$ and thus we can prune it without any damage.

But, if the regularizer $\mathcal{R}(W)$ is chosen to be L1 or Group Lasso, the $\epsilon$-closure $\{W| \mathcal{R}(W)<\epsilon\}$ would be the ball centered at zero.
Hence, minimizing $\mathcal{R}(W)$ cannot send $W$ to pruning-invariant set and hence inevitably induces pruning-caused damage on performance.
To achieve lossless pruning, we need to locate the parameter on $X_{tgt}$ by proper minimization process. Unfortunately, finding direction toward $X_{tgt}$ is difficult. Since $X_{tgt}$ is a union of linear subspaces, the exact defining equations of $X_{tgt}$ would be given by $N_W$ degree multivariate polynomials, which are not able to be efficiently computed or minimized. 

This difficulty of minimization is easily resolved by adding \emph{catalyst}, the additional parameters whose cardinality is $N_W$. With viewpoint of higher dimensional space, we can simplify the defining equation of $X_{tgt}$, by following theorem.

\begin{restatable}{theorem}{PruningConstraintThm}\label{thm:V(DW)-V(D)}
Let 
    \begin{equation}
        \mathbb{D}=\{D=diag(\delta)|\delta\in\mathbb{R}^{N_W}\}
    \end{equation}
be space of diagonal matrices and consider projection map 
    \begin{equation}
        \begin{aligned}
            p:\mathbb{R}^{N_W\times N_I}\times\mathbb{D}&\longrightarrow\mathbb{R}^{N_W\times N_I}\\
            (W,D)&\longmapsto W
        \end{aligned}
    \end{equation}
then 
\begin{enumerate}
    \item[(1)] $X_{tgt}=p(\{(W,D)|DW=0 \mbox{ and } D\neq 0\})$.
    \item[(2)] Let $B(X,\epsilon)$ be the $\epsilon$ neighborhood of $X$ with the $L_2$ norm. If $k>0$ is a positive real number, then \begin{equation}
        B(X_{tgt},\epsilon) = p(\{(W,D)|\|DW\|_{2,1}<k\epsilon \mbox{ and } \|D\|_1>k\})
    \end{equation}
\end{enumerate}
    
\end{restatable}

In this paper, we propose minimizing 
\begin{equation}
    \|DW\|_{2,1} = \sum_{i=1}^{N_W}\|(DW)_{i,:}\|_2 = \sum_{i=1}^{N_W} \|D_{ii}F_i\|_2,
\end{equation}
in extended parameter space as Catalyst regularization.
The minimization of $\|DW\|_{2,1}$ has good properties that
\begin{enumerate}
    \item The global minima of $\|DW\|_{2,1}$ is nontrivial and admits lossless pruning, because if $DW=0$ and $D\neq0$ then $W$ must lie in $X_{tgt}$. %Hence, lossless pruning is possible at the minima of $\|DW\|_{2,1}$.
    \item Although $\|DW\|_{2,1}$ is not convex, all critical points are global minima. Hence, we can reach the global minima by simple gradient descent.
    \item The dynamics of $\|DW\|_{2,1}$ minimization shows wide-margin bifurcation of magnitudes between the pruning filters and the preserving filters. This is described in \cref{subsec:bifurcation}.
\end{enumerate}

Using the catalyst $D$ and $\|DW\|_{2,1}$ requires a proper training algorithm that includes model extension and contraction methods. In \cref{subsec:bypassing} we present Catalyst Pruning algorithm. Interestingly, the constraint for lossless pruning $DW=0$ is also a constraint for the contraction to original parameter space. 

\subsection{bypassing algorithm for Catalyst pruning}\label{subsec:bypassing}

The algebraic constraint for the pruning is defined on extended parameter space with diagonal matrix $D$ in \cref{subsec:algebraic}. To exploit this constraint with external parameter $D$, we modify bypass algorithm\cite{bypass} to manage the model extension and regularization for the implementation.

Referring \cite{bypass}, the Bypass pipeline is comprised of the following three core components:
\begin{enumerate}
    \item $\varphi_2:\mathcal{D}_2\rightarrow\mathcal{F}$ : extension of model $\varphi_1$ satisfying $Im\varphi_1\subseteq Im\varphi_2$
    \item $embed$: extends the model parameter $\theta\in\mathcal{D}_1$ into extended space by 
    \begin{equation}
        embed:\mathcal{D}_1\rightarrow \mathcal{D}_2 \mbox{ s.t. } \varphi_1 = \varphi_2\circ embed
    \end{equation}
    \item $proj$ : contracts the model parameter $\omega\in\varphi_2^{-1}(Im\varphi_1)$ into original space by
    \begin{equation}
        \begin{aligned}
            proj: \varphi_2^{-1}(Im\varphi_1)\rightarrow \mathcal{D}_1 \mbox{ s.t. } \\
            \varphi_2 = \varphi_1\circ proj \mbox{ on } \varphi_2^{-1}(Im\varphi_1)
        \end{aligned}
    \end{equation}
\end{enumerate}

We adopt the Bypass pipeline as the implementation framework for our Catalyst pruning method, with a minor modification to the learnable activation function. Instead of the original form, which uses $\psi_D(x) = Dx + \sigma(x)$, we introduce a modified version:
\begin{equation}
    \psi_{D,\overline{D}}=Dx-\overline{D}x+\sigma(x)
\end{equation}
where $\sigma(x)$ denotes the nonlinear activation, to initialize the Catalyst $D$ with specific nonzero diagonal matrix $D^{init}$, which will be described in \cref{subsec:bifurcation}.

This modification aligns with our lossless pruning constraint. Specifically, the comeback constraint from the Bypass pipeline, given by $ADW = 0$, is automatically satisfied under our formulation, because our pruning condition enforces $DW = 0$. This allows us to simultaneously prune the regularized filters in $W$ and eliminate the auxiliary parameter $D$ after pruning.

We denote the composition of the pruning operation and the projection step as prune, with its full formulation provided in \cref{subsubsec:bypass-proj}.

\subsection{Bifurcation behavior of \texorpdfstring{$\|DW\|_{2,1}$}{DW norm} minimization}\label{subsec:bifurcation}
\begin{figure*}[t] % eggholder_projection
  \centering
    \begin{subfigure}[b]{0.33\textwidth}
    \centering
      % \includesvg[width=\textwidth]{plots/DW_dynamics/pruning_dw_dynamics.svg}
      \includegraphics[width=\textwidth]{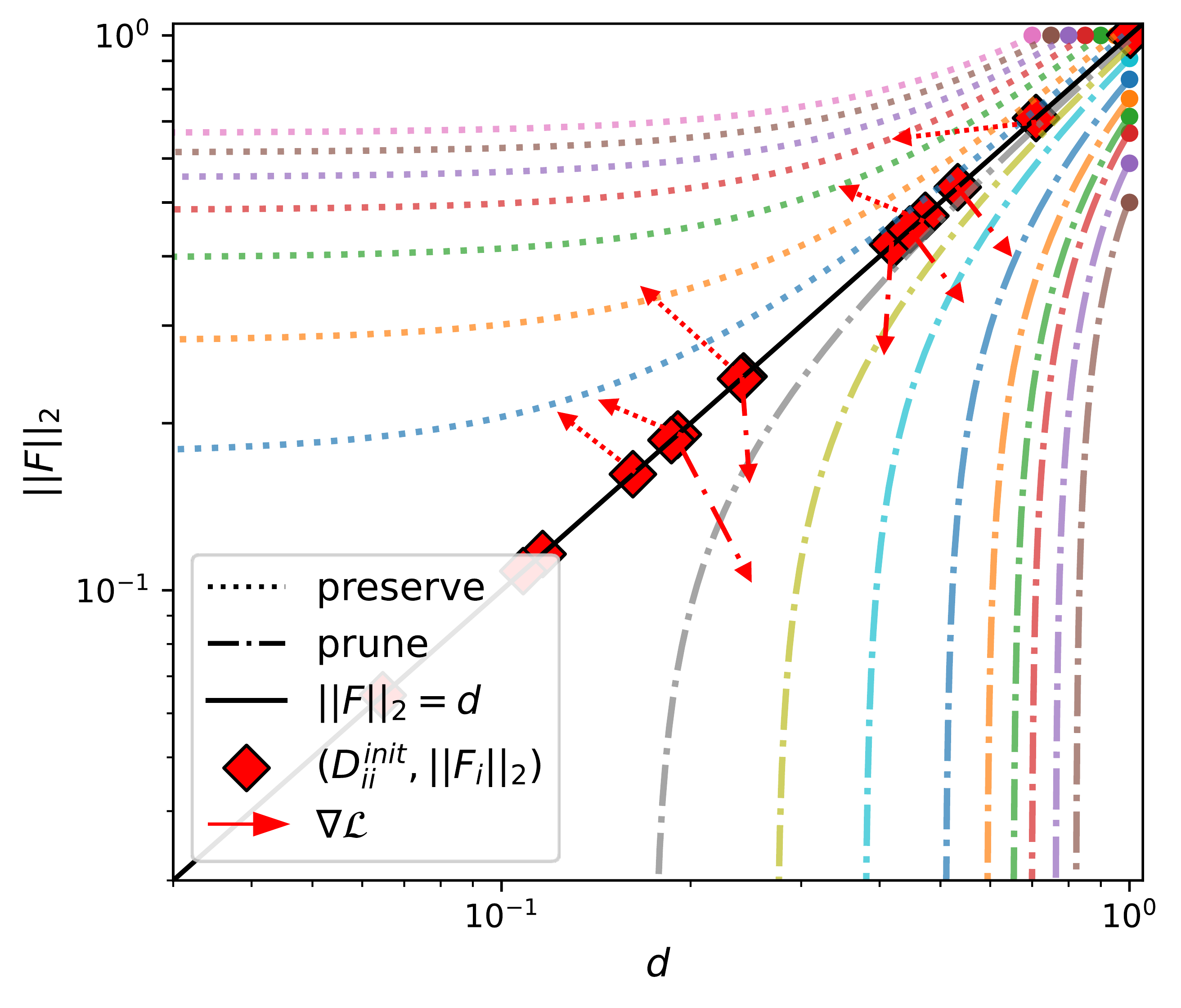}
      \caption{$(d,F)$ dynamics}
      \label{subfig:DW_simul_dynamics}
    \end{subfigure}%
    \hfill%
    \begin{subfigure}[b]{0.33\textwidth}
    \centering
      \includegraphics[width=\textwidth]{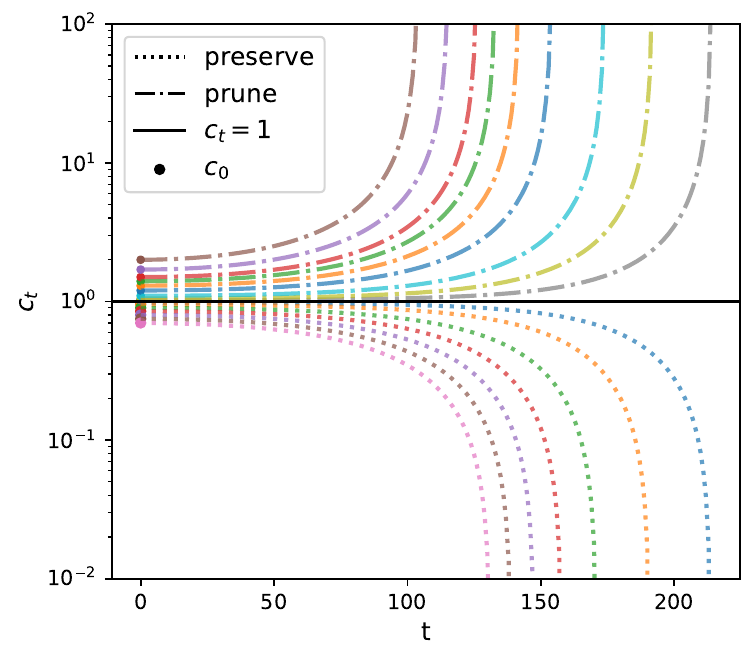}
      \caption{bifurcation on $c_t$}
      \label{subfig:DW_simul_dynamic_ct}
    \end{subfigure}%
    \hfill%
    \begin{subfigure}[b]{0.33\textwidth}
    \centering
      \includegraphics[width=\textwidth]{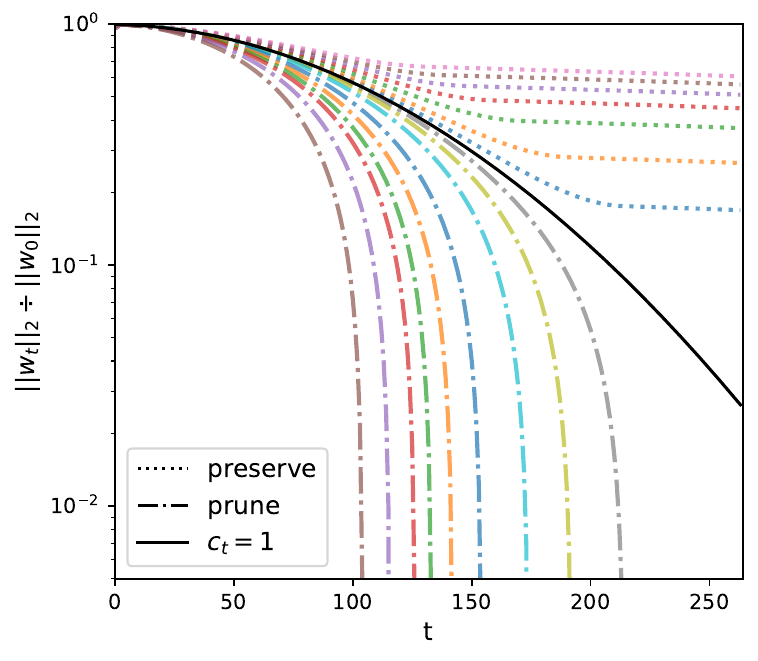}
      \caption{bifurcation on $W_t$}
      \label{subfig:DW_simul_dynamic_wt}
    \end{subfigure}%
\hfill%
\caption{Simulation on $\|DW\|_{2,1}$ minimization,}

\label{fig:DW_simulation}
\end{figure*}
In previous section, we proposed how to prune the model while minimizing $\|DW\|_{2,1}$. To complete the pipeline, we need to choose $D^{init}$ for $embed(D^{init})$. The choice of $D^{init}$ is crucial to control the sparsification, since the destination of $\|DW\|_{2,1}$ minimization is strongly dependent on the initial point $(D^{init},W)$.

In particular, for each filter $\{F_i\}_{i\in [N_W]}$ of $W$ we propose to set $D_{ii}^{init}$ to $c\|F_i\|_2$ with the hyperparameter $c=1$ where $N$ is the number of parameters in $F_i$. 
For all $i$, if $D_{ii}^{init}$ is set to $\|F_i\|_2$, then $(D_{ii}^{init},F_i)$ is located on the decision boundary (black line and red stars in \cref{subfig:DW_simul_dynamics}) on $(d,F)$-space, which is boundary of scaled $L_2$-cone and defined by equation
\begin{equation}
    \|F\|_2-d =0.
\end{equation}

During training, due to performance loss $\mathcal{L}$ (red arrows in \cref{subfig:DW_simul_dynamics}), the $(D_{ii},F_i)$ escapes this decision boundary almost surely, and then move according to its position whether the ratio $c_0^{(i)}=\frac{D_{ii}}{\|F_i\|_2}$ is larger than $1$ or not. This movement is illustrated in \cref{fig:DW_simulation} by dashed lines.
Precisely, according to \cref{thm:DW-dynamics} below, if $c_0^{(i)}>1$ then $i$th filter parameters $F_i$ the parameters $(D_{ii},F_i)$ move toward hyperplane $F=0$ of $(d,F)$-space and thus $i$th filter would be pruned. Otherwise if $c_0<1$, then the destination would be $d=0$ and the $i$th filter would be preserved.

\begin{restatable}{theorem}{DWdynamicsThm}\label{thm:DW-dynamics}

    Let $d$ be the scalar parameter and $M$ be N-dimensional vector. $M_t\mbox{ and }d_t$ be the trajectory of gradient descent movement of $\|dM\|_2$ at timestep $t$ positive learning rate $\lambda_t$ and weight decay term $\alpha \ll 1$. Let $c_t=\frac{|d_t|}{\|M_t\|_2}$ and assume that 
    \begin{equation}\label{eqn:lambda_t_cond_c0}
        0<\lambda_* < \lambda_t \ll  \min\left( \frac{(1-\alpha)}{c_0},(1-\alpha)c_0 \right)
    \end{equation}
    where $\lambda_*=\inf_{t}\lambda_t$ is the infimum $\lambda_t$. 
    \begin{enumerate}
        \item[(1)] If $c_0=1$, then $c_t=1$ for all timestep $t$.
        \item[(2)] If $c_0<1$, then $c_t$ exponentially shrinks to $\frac{\lambda_t}{1-\alpha}$. 
            i.e, If T be the smallest integer satisfying $c_T \leq \frac{\lambda_T}{1-\alpha}$ 
            then there exists $k\in (0,1)$ such that $c_{t+1}\leq kc_t$ for all $t<T-1$.
        \item[(3)] If $c_0>1$, then $c_t$ exponentially grows to $\frac{(1-\alpha)}{\lambda_t}$.
        i.e, if T is the smallest integer satisfying $c_T\geq \frac{1-\alpha}{\lambda_T}$
        then there exists $k>1$ such that $c_{t+1}\geq kc_t$ for all $t<T-1$
    \end{enumerate}
\end{restatable}
\begin{proof}[Proof of \cref{thm:DW-dynamics}]
    While the sign of each entries of $M$ or d is unchanged, we can find recurrence relation between $c_t$ and $c_{t+1}$. We can prove that multiplication coefficient of recurrence relation depends on the initial value $c_0$ completely, and thus $c_t$ exponentially grows or shrinks. The detailed proofs can be found in \cref{appendix:DWdynamicsThm}.
\end{proof}

The bifurcation behavior explained by \cref{thm:DW-dynamics} is empirically observed in the example dynamics plot in \cref{fig:DW_simulation}, that bifurcation on $c_t$ in \cref{subfig:DW_simul_dynamic_ct} results in the bifurcation between two classes of filters which would be pruned or preserved.
This gives rise to a very effective and robust decision strategy to decide which filters would be pruned. As ratio $c_t=\frac{D_{ii}}{\|F_i\|_2}$ bifurcates to getting away from $1$, we may prune the $i$th filter when $\frac{D_{ii}}{\|F_i\|_2}>1$ and construct a set of pruning indices $P=\{i| D_{ii}>\|F_i\|_2\}$. 

Since the pruning decision of $i$th filter is made by the position of $(D_{ii},F_{i})$ induced by performance loss $\mathcal{L}$, all channels get provably equal chance to be pruned. That is, filters with smaller initial magnitudes are not any more preferred to be pruned than those with larger magnitudes.

\section{Implementation and Empirical Validation}\label{sec:experiments}

\begin{table*}[t]
    \centering
    \caption{Performance Comparison of Various Filter Pruning Methods}
    \resizebox{0.9\linewidth}{!}{%
    \begin{tabular}{cccccc}
\toprule
\multirow{2}{*}{} & \multirow{2}{*}{Method} & Baseline & Pruned & \multirow{2}{*}{$\Delta$ ACC(\%)} & \multirow{2}{*}{speedup} \\
                                   &                                                   & ACC(\%) & ACC(\%) &       &        \\ \midrule
\multirow{7}{*}{Resnet56+ CIFAR10} & Slimming\cite{liu2017slim,zhuang2020polarization} & 93.80   & 93.27   & -0.53 & 1.92x  \\
                                   & Polar\cite{zhuang2020polarization}                & 93.80   & 93.83   & +0.03 & 1.89x  \\
                                   & SCP \cite{kang2020scp}                            & 93.69   & 93.23   & -0.46 & 2.06x  \\
                                   & ResRep\cite{ding2021resrep}                       & 93.71   & 93.71   & 0.00  & 2.12x  \\
                                   & Depgraph\cite{fang2023depgraph}                   & 93.53   & 93.77   & +0.24 & 2.13x  \\
                                   & ATO\cite{wu2024ato}                               & 93.50   & 93.74   & +0.24 & 2.22x  \\
                                   & Ours-BN                                           & 93.53   & 94.00   & +0.47 & 2.06x  \\ \midrule
\multirow{11}{*}{VGG19+ CIFAR100}  & SCP\cite{kang2020scp}                             & 72.56   & 72.15   & -0.41 & 2.63x  \\
                                   & Greg-1 \cite{wang2020greg}                        & 74.02   & 71.30   & -2.72 & 2.96x  \\
                                   & DepGraph\cite{fang2023depgraph}                   & 73.50   & 72.46   & -1.04 & 3.00x  \\
                                   & Ours-BN                                           & 73.51   & 73.37   & -0.14 & 3.00x  \\ \cline{2-6} 
                                   & DepGraph\cite{fang2023depgraph}                   & 73.50   & 70.39   & -3.11 & 8.92x  \\
                                   & Greg-2 \cite{wang2020greg}                        & 74.02   & 67.75   & -6.27 & 8.84x  \\
                                   & EigenD\cite{wang2019eigendamage}                  & 73.34   & 65.18   & -8.16 & 8.80x  \\
                                   & Ours-BN                                           & 73.51   & 70.08   & -3.43 & 8.96x  \\ \cline{2-6} 
                                   & DepGraph\cite{fang2023depgraph}                   & 73.50   & 66.20   & -7.30 & 12.00x \\
                                   & Ours-BN                                           & 73.51   & 68.13   & -5.38 & 11.84x \\
                                   & Ours-BN                                           & 73.51   & 66.44   & -7.07 & 13.83x \\ \midrule
\multirow{11}{*}{Resnet50+Imagenet} & PFP\cite{Liebenwein2020Provable} & 76.13 & 75.21 & -0.91 & 1.49x \\
                                    &Greg-1\cite{wang2020greg}& 76.13& 76.27 & +0.14 & 1.49x \\
                                    &Ours-BN& 76.15& 76.40 & +0.36 & 1.49x \\ \cline{2-6}
                                    &ThiNet70\cite{luo2017thinet} & 72.88 & 72.04 & -0.84 & 1.69x \\
                                    &Whitebox\cite{whitebox} & 76.15 & 75.32 & -0.83 & 1.85x \\
                                    &Ours-BN & 76.15 & 76.04 & -0.11 & 1.82x \\ \cline{2-6}
                                    &Slimming\cite{liu2017slim,zhuang2020polarization} & 76.15 & 74.88 & -1.27 & 2.13x \\
                                    &Polar\cite{zhuang2020polarization} & 76.15 & 75.63 & -0.52 & 2.17x \\
                                    &Depgraph\cite{fang2023depgraph} & 76.15 & 75.83 & -0.32 & 2.08x \\
                                    &OICSR\cite{li2019oicsr} & 76.31 & 75.95 & -0.37 & 2.00x \\
                                    &Ours-group & 76.15 & 75.96 & -0.19 & 1.96x \\ \bottomrule
\end{tabular}
}
    \label{tab:performance}
\end{table*}

\subsection{Pruning Algorithm: Catalyst pruning}\label{sec:algorithm}

The proposed algorithm starts with $embed(D^{init})$, where $D^{init}$ is set to be $D^{init}= c\times diag(\|F_1\|_2,\cdots,\|F_{N_W}\|_2)$ with $c=1$, as proposed in \cref{subsec:bifurcation}. The $c=1$ is proposed to place the pair of $(D_{ii},F_i)$ on the pruning decision boundary, but the practitioners may set this value to $c>1$ to prune more, or $c<1$ to prune less.

After initialization, the proposed algorithm repeats regularize-and-prune loop twice, to remove $D$ and $\overline{D}$ with pruning operation, respectively. 
During the first loop, namely $opt_1$, we minimize $\mathcal{L}+\gamma_t(\|DW\|_{2,1})$ with SGD optimizer until the training budget $T$. The $\gamma_t$ is the parameter which controls the weight of regularization as in \cite{bypass}. If $\|DW\|_{2,1}$ decreases to small positive value $\epsilon$ or the $c_t$ are bifurcated enough to satisfy $|log (c_t)|>\kappa$ for all channels, the regularization loop of $opt_1$ may be stopped early.
After regularization stage, we choose the pruning indices by $P=\{i|D_{ii}>\|F_i\|_2\}$, and prune the selected filters by $prune(P)$ defined in \cref{subsec:bypassing} to obtain intermediate pruning results with extra (but pruned) parameter $\overline{D}$. 
Applying similar loop, but with $\overline{D}=0$, we can prune the model again and obtain pruned model with original architecture. We summarize the proposed algorithm as a pseudo code in \cref{appendix:algorithm}.

\subsection{Models, Datasets and Settings}\label{subsec:exp-setting}

We use Resnet-56 \cite{xie2017resnet} on CIFAR10 \cite{krizhevsky2009cifar}, VGG-19 \cite{simonyan2014vgg} on CIFAR100 and Resnet50 on Imagenet \cite{ILSVRC15_imagenet} for empirical verification. 
The detailed experimental settings including augmentation and hyperparameters such as learning rate, number of epochs, $\epsilon$ and $T$ can be found in \cref{appendix:hyperparameters}.

Catalyst pruning is applicable to various type of layers such as convolution, fully connected layers, or even grouped multiple layers, as \cref{thm:DW-dynamics} does not limit the dimension of the regularization target dimension.

In this work, we consider two pruning targets: the scaling factors of BN and grouped parameters\cite{fang2023depgraph}, which are denoted by \textbf{Ours-BN} and \textbf{Ours-group} in  each table. The $\sigma$ would be the (ReLU) activation of each block if the target is BN, otherwise $\sigma$ would be the whole BN layer.

For the Imagenet experiment, to show only the effect of our method, we follow the strict principle that no extra engineering skills such as additional augmentation, label smoothing, mixup, etc. are used, but only the techniques used for the official recipe for the pretraining, provided by TorchVision\cite{torchvision2016}. 
The benchmark models using extra skills (and compared to models trained with vanilla settings) were excluded from \cref{tab:performance}.

\subsection{Pruning Performance Comparison}\label{subsec:res-performance}

We present the pruning performance of tested filter pruning methods \cite{liu2017slim,zhuang2020polarization,kang2020scp,wang2020greg,luo2017thinet,wu2024ato,fang2023depgraph,li2019oicsr,wang2019eigendamage,ding2021resrep} in \cref{tab:performance}. %-cifar10},\cref{tab:prerformance-cifar100} and \cref{tab:performance-imagenet}. 
The results with similar pruning ratio or similar speedups are partitioned with horizontal rules, and it is evident that our proposed method achieves promising performance in most partitions. 
Compared to batch normalization layer regularization methods \cite{liu2017slim,zhuang2020polarization,kang2020scp}, our method shows superior performance that allows to maintain the accuracy during the pruning. 

\subsection{Robust Bifurcation}\label{subsec:res-polarization}

\begin{figure}[t]
  \centering
    \begin{subfigure}[b]{0.33\textwidth}
    \centering
      \includegraphics[width=\textwidth]{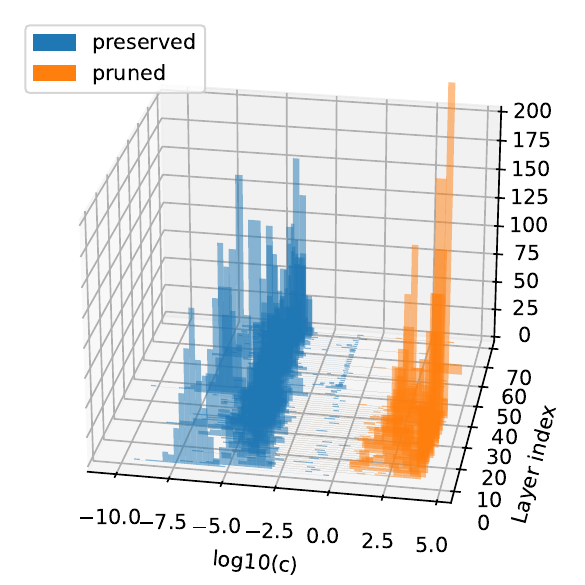}
      \caption{Bifurcation on $c$}
      \label{subfig:polar_C}
    \end{subfigure}%
    \hfill%
    \begin{subfigure}[b]{0.33\textwidth}
    \centering
      \includegraphics[width=\textwidth]{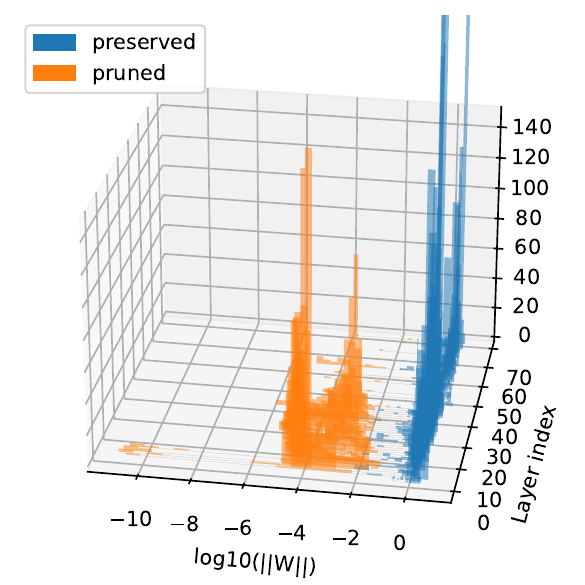}
      \caption{Bifurcation on $W$}
      \label{subfig:polar_W}
    \end{subfigure}%
    \hfill%
    \begin{subfigure}[b]{0.33\textwidth}
    \centering
      \includegraphics[width=\textwidth]{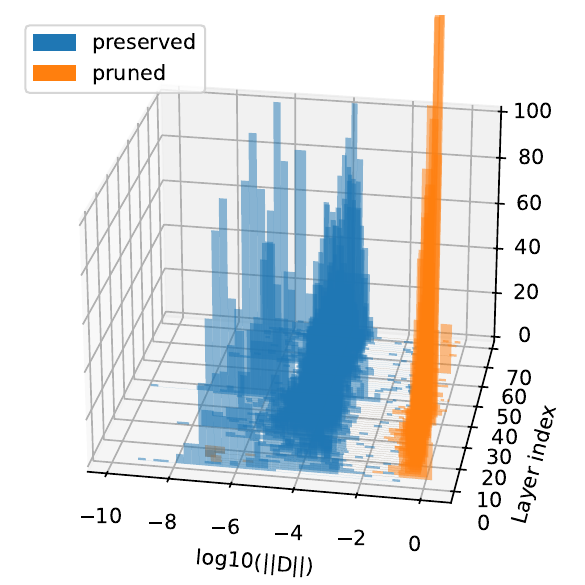}
      \caption{Bifurcation on $D$}
      \label{subfig:polar_D}
    \end{subfigure}%
\hfill%
\caption{The histograms of ratio $c=\frac{D_{ii}}{\|F_i\|}$, filter vector $F_i$ of weight tensor $W$, and $D_{ii}$'s for each layers of Resnet50-1.8G model trained on Imagenet. The z-axis represents the frequency. The layer indices of results in $opt2$ are shifted to start from the last layer.}
\label{fig:polarization}
\end{figure}

We present a typically observed bifurcation behavior catalyst regularization in \cref{fig:polarization}. 
\cref{subfig:polar_C} shows robust bifurcation of $c=\frac{D_{ii}}{\|F_i\|}$ centered at $c=1$, by $\|DW\|_{2,1}$ minimization, as theoretically expected in \cref{thm:DW-dynamics}. The ratio of $c$ between pruned filters and preserved filters are extremely large, around $10^8$, since the $c$ value changes exponentially during the minimization.
The bifurcation on $c$ naturally induces the bifurcation on $W$, the filters on the target layer, due to its definition. 
Evidence of bifurcation on the  $W$ bifurcation is visualized in \cref{subfig:polar_W} that the filter norms of pruned filters are pushed to have small value, while the filter norms the rest are preserved.
This property is not observed in conventional regularization methods since they push all filters to zero. This property can be observed in \cref{appendix: plots_init_magnitude_preference}, that the filter norm of regularized filters are reduced in 1st and 2nd columns of each plot.
The bifurcation on $D$ was also observed, that the $D$ of preserved filters sacrificed instead of $W$, and the $D$ value of pruned filters are preserved to be large. 

Since our pruning decision is based on $c$, our algorithm can make not-to-prune decision when the loss gradient is likely to preserve the filter norm. 
In \cref{fig:polarization}, some layers were not pruned in $opt_2$ phase since all $\|F_i\|$ were preserved and $D_{ii}$ are all pushed toward zero instead.

We also present an all-layer bifurcation comparison for VGG19 on CIFAR100 in the bottom row figures in \cref{appendix: plots_init_magnitude_preference}, in which our catalyst regularization shows robust bifurcation whereas $L_1$ and Group Lasso do not.
Also, the top row figures in \cref{appendix: plots_init_magnitude_preference} contain empirical evidence on Catalyst pruning algorithm providing fair pruning chance, whereas $L_1$ and Group Lasso regularization tend to prune filters with smaller magnitudes.
\subsection{Empirical Evidence of Lossless Pruning}\label{subsec:res-equivalence}

\noindent

The proposed catalyst regularization is designed to be lossless, which refers no performance damage at the pruning operation. 

\begin{minipage}[t]{0.48\textwidth}
  \centering
\captionof{table}{Effect of $prune$ operation in Catalyst pruning.}
\begin{adjustbox}{width=\linewidth}
\begin{tabular}{cc@{\hskip3pt}c@{\hskip3pt}cc}
\toprule
 \multirow{2}{1.6cm}{\hfil Task}& \multirow{2}{*}{phase} & \multirow{1}{*}{\hfil avg} & \multirow{1}{*}{\hfil  avg}&MACs \\
 &  & $\Delta$ acc (\%p) & $\Delta \mathcal{L}$ (\%)&drop(\%) \\
\midrule
\multirow{2}{1.6cm}{\hfil R56\_2.06x CIFAR10} & $opt_1$ & -0.001&0.0032&9.062\\
 & $opt_2$ &0.004&-0.0025&46.562\\
 \midrule
\multirow{2}{1.6cm}{\hfil V19\_8.96x CIFAR100} & $opt_1$ & -0.019&0.0001&85.075\\
&$opt_2$&0.009&0.0021&25.179\\
 \midrule
\multirow{2}{1.6cm}{\hfil R50\_1.82x Imagenet} & $opt_1$ & 0.002&-0.0001&19.233\\
&$opt_2$&-0.037&0.0008&31.811\\
 \bottomrule
\end{tabular}
\end{adjustbox}
  \label{tab:lossless}
\end{minipage}\hfill
\begin{minipage}[t]{0.48\textwidth}
\vspace{0.1em}
  \centering
  \includegraphics[width=\linewidth]{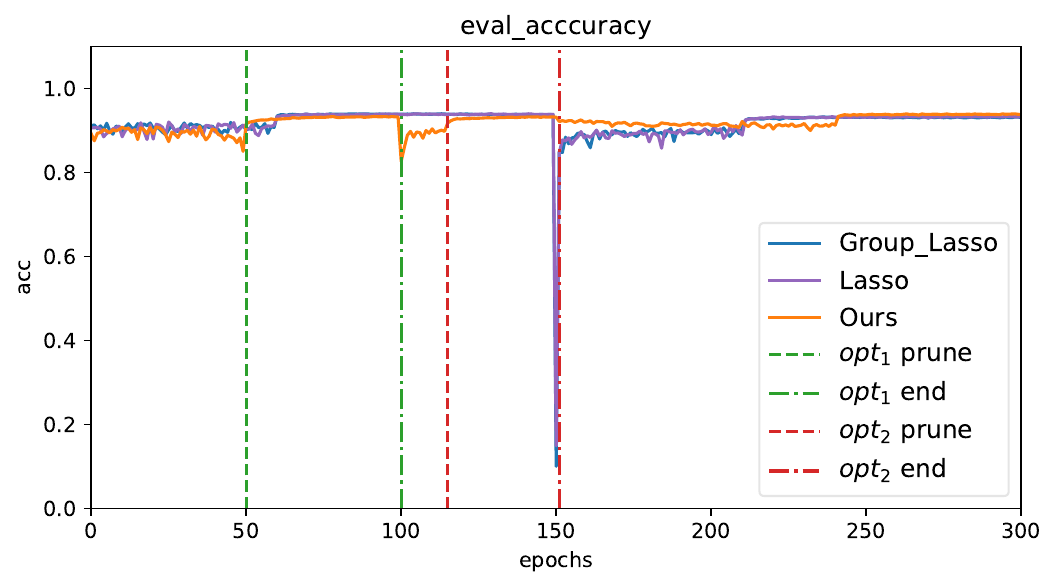}
  \captionof{figure}{Accuracy graph of Resnet56+CIFAR10 training, compared to Lasso and Group Lasso regularization.}
  \label{fig:acc_curve_lasso}
\end{minipage}

To check the proposed pruning operations were lossless, we report the average of test accuracy drop and test loss difference for each pruning operations in $opt_1$ and $opt_2$ in \cref{tab:lossless} with total MACs drop of each phase. 
Pruning at $\|DW\|_{2,1}<\epsilon$ shows promising results of lossless pruning, that the performance drop caused by $prune$ is negligible but also improves the performance. 
For CIFAR experiments with Resnet56 and VGG19, almost all layers except one or two layers were pruned at same epoch (within few training steps), which is right after the learning rate decay. 
Unlike other regularization methods which often shows remarkable performance drop (e.g. Fig. 4 of \cite{ding2021resrep}) if the layers are pruned together after the regularization, the Catalyst pruning shows very mild temporary loss during the entire pruning procedure, as shown in \cref{fig:acc_curve_lasso} and \cref{fig:loss_curve} in \cref{sec:loss_curves_during}.
To the best of our knowledge, pruning methods with such stable loss trajectory has not been reported in the structured pruning literature.

\section{Discussion}\label{sec:discussion}

To present a compression method rather than regeneration, we argue that intermediate performance of regularization-based pruning method should be preserved properly throughout the process, rather than relying on recovery after substantial degradation.
The Catalyst pruning shows lossless pruning after regularization, and thus provides the smooth learning curve,
but however the training-caused damage is observed.
This training-caused damage is inevitable in constrained optimization yet can be readily ameliorated by taking smaller $\gamma_t$ with more training budget, similarly to $opt_2$ of \cite{bypass}. 
Alternatively, we may bring other constrained optimization algorithm for the improvement.

Unlike conventional Lasso regularizers, the pruning decision of the Catalyst pruning is independent from the magnitude of the pretrained model.
The decision of whether a given filter is pruned depends on whether the ratio $c=\frac{D_{ii}}{\|F_i\|_2}$ falls below or above $1$ during the training process, which follows the gradient of performance loss $\mathcal{L}$.
The ratio $c$ has geometric meaning in the extended space, when we define projective space on it, that it is in fact a cotangent of angular distance between representation or zero-filter and current parameter.
As a future work, we will investigate this intuition in projective geometry for another application.

\section{Conclusion}\label{sec:conclusion}

We propose an algebraically principled definition of lossless structured pruning, and use it as a blueprint to design a novel regularizer with provable potential to achieve lossless pruning. 
We name the regularizer Catalyst, as it acts like a catalyst for pruning such that the to-be-pruned neural network is temporarily deformed prior to actual pruning operations. 
We use Catalyst to construct a novel structured pruning algorithm, named Catalyst pruning, which enjoys provable zero-bias fair-chance pruning behavior and robust pruning decision boundary with wide-margin bifurcation dynamics due to theoretical guarantees naturally arising from the Catalyst regularizer. 
Empirical validations support the theoretically expected benefits of Catalyst regularizer, as Catalyst pruning shows superior performances across various benchmarks when compared to filter pruning methods with conventional regularizers.

{
    \small
    \bibliographystyle{unsrt}
    \bibliography{jaeheun}
}

\appendix
\onecolumn
\counterwithin*{equation}{section}
\renewcommand\theequation{\thesection\arabic{equation}}

\section{Components of \cref{subsec:bypassing}}
In this section, we provide detailed explanations on each components of Bypassing, focusing on the modifications for Catalyst pruning. 

\subsection{Model extension}\label{sub^2sec:bypassing_pruner}

Unlike the original bypass algorithm \cite{bypass}, we propose to use learnable activation which employs two parameters $D$ and $\overline{D}$ for extended model $\varphi_2$, to give nontrivial initialization on $D$, as follows:

\begin{definition}\label{def:extension}
    Let $\overline{\theta} = (W,b_W,A,b_A)$ and rewrite 
    \begin{equation}
        \overline{\varphi_1}(\overline{\theta}) = \overline{\varphi_1}(W,b_W,A,b_A)=NN(W,b_W,A,b_A,\sigma).  
    \end{equation}
    \begin{enumerate}
        \item We first define learnable activation $\psi_{D,\overline{D}}$ with additional parameter $D$ and $\overline{D}$:
        \begin{equation}
            \psi_{D,\overline{D}}:x\mapsto Dx-\overline{D}x+\sigma(x)
        \end{equation}
        \item We define $\overline{\varphi_2}$ by
        \begin{equation}
        \begin{aligned}
            \overline{\varphi_2}(\overline{\theta},D,\overline{D}) &= \overline{\varphi_2}(W,b_W,A,b_A,D,\overline{D}) \\
            &= NN(W,b_W,A,b_A,\psi_{D,\overline{D}})
        \end{aligned}
        \end{equation}
        \item The $\varphi_2(\theta,D,\overline{D})$ is defined by replacing $\overline{\varphi_1}(\overline{\theta})$ to $\overline{\varphi_2}(\overline{\theta},D,\overline{D})$ from original model $\varphi_1(\theta)$.
    \end{enumerate}    
\end{definition}

\subsection{The \emph{embed} function}

Now we set the $embed$ map as follows:
\begin{definition}\label{def:embed}
Given $D^{init}$, we define $embed(D^{init})$ as function-preserving operator on extended parameter space, as follows:
\begin{equation}
    embed(D^{init}): \overline{\theta}\mapsto (\overline{\theta},D^{init},D^{init})
\end{equation}
\end{definition}

\begin{remark}\label{rmk:embed}
The map $embed$ defined in \cref{def:embed} has a function-preserving property on $\varphi_2$, that $\varphi_2=\varphi_2\circ embed(D^{init})$ because $\psi_{D^{init},D^{init}} = \sigma$.
\end{remark}

For the implementation, we need to choose $D^{init}$ to complete the model extension. In this paper, we proposed to use $D^{init} = diag(\|F_1\|_2,\cdots,\|F_{N_W}\|_2)$, where $N$ is the number of entries in each filter. 
With the proposed initialization, the fair pruning chance would be ensured and the $\|DW\|_{2,1}$ minimization would show bifurcation behavior. The detailed explanation and mathematical backgrounds which supports this proposed initialization are included in \cref{subsec:bifurcation} with simulations.

\subsection{\emph{prune}: combination of pruning and \emph{proj}}\label{subsubsec:bypass-proj}
Now we propose the last core component of Catalyst pruning by defining contraction map $prune$, which replaces $proj$ of bypass pipeline.
\begin{definition}\label{def:prune}
Let $P$ be set of filter indices to be pruned and $P^c$ be its complement. 

We define $prune$ by
        \begin{equation}
        \begin{aligned}
            &prune(P):(W,b_W,A,b_A,D,\overline{D})\mapsto\\
            &(W[P^c],b_W[P^c],(A^T[P^c])^T, b_A',-\overline{D}[P^c],0)
        \end{aligned}
        \end{equation}
        where $b_A'=b_A+ADb_W+(A^T[P])^T(\psi_{-\overline{D},0}(b_W))[P]$
        
\end{definition}

\begin{restatable}{theorem}{ProjThm}\label{thm:projection}
    Let $prune$ be the mappings defined in \cref{def:prune}. If $DW=0$ and $P=\{i|D_{ii}\neq 0\}=\{i|\forall j W_{ij}=0\}\in 2^{N_W}$, then $prune(P)$ becomes function-preserving. That is, 
    \begin{equation}
        \overline{\varphi_2} = \overline{\varphi_2}\circ prune(P).
    \end{equation}
\end{restatable}
\begin{proof}[Proof of \cref{thm:projection}]
   For arbitrary input, if $DW=0$ then some filters would provide constant features. We pass those constant filters to next layer and add them to bias vector of next layer. The detailed proof is presented in \cref{appendix:ProjThm}.
\end{proof}

Combining $\varphi_2$, $embed(D^{init})$ and $prune(P)$ we can build the training pipeline for the structured pruning, with constrained optimization algorithm, $opt_1$ and $opt_2$. 
Starting with $\varphi_1(\theta)$, we extend the model to $\varphi_2(\theta,D,\overline{D})$ by $embed(D^{init})$, train with constrained optimization $opt_1$ and algebraic constraints $DW=0$, and prune the model by $prune(P)$ to get pruned but still extended model $\varphi_2(\theta[P],D[P],0)$. Again, with constrained optimization $opt_2$ and $\|DW\|_{2,1}=0$, we get final pruned model by second $prune$. 
As stated in \cref{rmk:embed} and \cref{thm:projection}, the $embed(D^{init})$ and $prune(P)$ are function-preserving operations and thus there would be no pruning-caused damage if $\|DW\|_{2,1}=0$.

\section{Pseudocode of Catalysis pruining}\label{appendix:algorithm}
We present our implementation of Catalyst pruning in \cref{alg:mildpruning} which is adaptation of Bypass pipeline \cite{bypass} with catalyst regularization and other theoretical aspects described in \cref{sec:sparsity_reg}. 
The algorithm can be summarized as: extend the model $\varphi_1$ by introducing additional parameters $D$ and $\overline{D}$, train in extended space with constrained optimization, and contract back to the original model.

The proposed algorithm starts with $embed(D^{init})$, where $D^{init}$ is set to be $D^{init}= c\times diag(\|F_1\|_2,\cdots,\|F_{N_W}\|_2)$ with $c=1$, as proposed in \cref{subsec:bifurcation}. The $c=1$ is proposed to place the pair of $(D_{ii},F_i)$ on the pruning decision boundary, but the practitioners may set this value to $c>1$ to prune more, or $c<1$ to prune less.

After initialization, the proposed algorithm repeats regularize-and-prune loop twice, to remove $D$ and $\overline{D}$ with pruning operation, respectively. 
During the first loop, namely $opt_1$ (line 3-6 in \cref{alg:mildpruning}), we minimize $\mathcal{L_2(\overline{\theta},D,\overline{D})}+\gamma_t(\|DW\|_{2,1})$ with SGD optimizer until the training budget $T$. The $Optimizer_\lambda(\cdot)$ represents the single SGD update with hyperparameter $\lambda$ and
$\gamma_t$ is the parameter which controls the weight of regularization as in \cite{bypass}. 

We can control the pressure of sparsification during $opt_1$, by changing $\alpha_\theta$ and $\alpha_D$ which are the weight decay terms of $\theta$ and $D$ each. Those are considered to be same in \cref{subsec:bifurcation}, but we may set $\alpha_\theta>\alpha_D$ to promote larger pruning ratio. In case of $\alpha_\theta>\alpha_D$, larger $\gamma_t$ would induce larger pressure on sparsification since the influence of $\nabla\mathcal{L}_2$ is weakened, compared to the deterministic movement of $\nabla \|DW\|_{2,1}$. 

If $\|DW\|_{2,1}$ decreases to small positive value $\epsilon$ (line 6 of \cref{alg:mildpruning}) or all $c_t$ are bifurcated enough to satisfy $|log (c_t)|>\kappa$, the regularization loop of $opt_1$ may be stopped early. For experiments,we used $\kappa=1$ for Imagenet and $\kappa=\infty$ for CIFAR, since early stopping was not necessary. 
After regularization stage, we choose the pruning indices by threshold $c=1$, which is equivalent to $P=\{i|D_{ii}>\|F_i\|_2\}$, and prune the selected filters by $embed(P)$ defined in \cref{subsec:bypassing} to obtain intermediate pruning results with extra (but pruned) parameter $D$. 
Applying similar loop in line 9-14 of \cref{alg:mildpruning}, but with $\overline{D}=0$, we can prune the model again and obtain pruned model with original architecture.
\begin{algorithm}[H]\label{alg:mildpruning}
    \caption{Regularization with catalyst}
    \begin{algorithmic}[1]
        \STATE \textbf{Input} $\overline{\theta}=(W,b_W,A,b_A),\lambda,\lambda',\mathcal{L}_2,\epsilon,\epsilon',\gamma_t,\gamma_t',T,T',c=1$
        \STATE \textbf{Initialize} $\overline{\theta},D,\overline{D}\leftarrow embed(D^{init})(\overline{\theta},0,0) \mbox{ where } D^{init}= c\cdot diag(\|F_1\|_2,\cdots,\|F_{N_W}\|_2)$ 

        \REPEAT
            \STATE $\overline{\theta},D,\overline{D} \leftarrow \mathrm{Optimizer}_\lambda(\mathcal{L}_2(\overline{\theta},D,\overline{D})+\gamma_t\|DW\|_{2,1})$
            \STATE $t\leftarrow t+1$
        \UNTIL{$\|DW\|_{2,1}<\epsilon$ or $log(c_t)>\kappa$ or $t>T$}
        \STATE $P\leftarrow \{i|D_{ii}>\|F_i\|_2\}$ 
        \STATE $\overline{\theta},D,0\leftarrow prune(P)(\overline{\theta},D,\overline{D})$
        \REPEAT
            \STATE $\overline{\theta},D,0 \leftarrow \mathrm{Optimizer}_{\lambda'}(\mathcal{L}_2(\overline{\theta},D,0)+\gamma_t'\|DW\|_{2,1})$
            \STATE $t\leftarrow t+1$
        \UNTIL{$\|DW\|_{2,1}<\epsilon'$ or $log(c_t)>1$or $t>T'$}
        \STATE $P\leftarrow \{i|D_{ii}>\|F_i\|_2\}$ 
        \STATE $\overline{\theta},0,0\leftarrow prune(P)(\overline{\theta},D,0)$
        \RETURN the pruned parameter $\overline{\theta}$ (and continue finetune.)
    \end{algorithmic}
\end{algorithm}

\section{Proof of \cref{thm:V(DW)-V(D)}}\label{appendix:pf_PruningConstraintThm}
\PruningConstraintThm*
    \begin{proof}[Proof of \cref{thm:V(DW)-V(D)} (1)]
        We first prove that $X_{tgt}\subseteq p(\{(W,D)|DW=0 \mbox{ and } D\neq 0\})$.

        if $\overline{W}\in X_{tgt}$, then there exists $\overline{i}$ such that $\overline{W_{\overline{i}}}=0$.
        Without loss of generality, let $\overline{i}=1$ and consider
        \begin{equation}
            \overline{D} = diag(1,0,\cdots,0).
        \end{equation}
        Then $\overline{D}\overline{W}=0$  and $\overline{D}\neq 0$.
        Therefore, we have 
        \begin{equation}
            (\overline{W},\overline{D}) \in \{(W,D)|DW=0 \mbox{ and } D\neq 0\}
        \end{equation}
        and hence 
        \begin{equation}
            \overline{W}=p(\overline{W},\overline{D})\in p(\{(W,D)|DW=0 \mbox{ and } D\neq 0\}).
        \end{equation}

        For the opposite inclusion, let $\tilde{W}\in p(\{(W,D)|DW=0 \mbox{ and } D\neq 0\})$.
        Then, there exist $\tilde{D}\neq 0$ that satisfy $\tilde{D}\tilde{W}=0$.

        Without loss of generality, we have $\tilde{D}_{11}\neq 0$ and thus $\tilde{W}_1=0$ since $(\tilde{D}\tilde{W})_1=0$. Therefore, $\tilde{W}$ in $X_{tgt}$ and \begin{equation}
            X=p(\{(W,D)|DW=0 \mbox{ and } D\neq 0\}).
        \end{equation}
    \end{proof}
    \begin{proof}[Proof of \cref{thm:V(DW)-V(D)} (2)]

        Suppose $W\in B(X_{tgt},\epsilon)$ and let $\overline{i} = argmin_{i\in[N_W]}\|W_{i,:}\|_2$. WLOG, let $\overline{i}=1$ then we have $\|W_{1,:}\|_2<\epsilon$ and thus we can choose $k'\in (k,\frac{\epsilon}{\|W_{1,:}\|_2}k)$.
        
         Let $\overline{D}=diag(k',0,\cdots,0)$. Then we have $\|D\|_1 = k'>k$ and 
        \begin{equation}
            \|DW\|_{2,1} = k'\|W_{1,:}\|_2+\sum_{j>1}0\cdot \|W_{j,:}\|_2<k\epsilon.
        \end{equation}
        Therefore, $W\in \{(W,D)|\|DW\|_{2,1}<k\epsilon \textrm{ and } \|D\|_1>k\}$.
        
        For the opposite direction, let $\|\tilde{D}\tilde{W}\|_{2,1}<k\epsilon$ and $\|\tilde{D}\|_1>k$. Let $\tilde{i}=argmin_{i\in[N_W]}\|\tilde{W}_{i,:}\|_2$ Then we have
        \begin{equation}
            \|\tilde{W}_{\overline{i},:}\|_2=\sum_{i=1}^{N_W}\frac{|\tilde{D}_{ii}|}{\|\tilde{D}\|_1}\|\tilde{W}_{\overline{i},:}\|_2\leq \sum_{i=1}^{N_W}\frac{|\tilde{D}_{ii}|}{\|\tilde{D}\|_1}\|\tilde{W}_{i,:}\|_2<
            \frac{1}{k}\sum_{i=1}^{N_W}|D_{ii}|\|\tilde{W}_{i,:}\|_2 = \frac{1}{k}\|\tilde{D}\tilde{W}\|_{2,1}<\epsilon
        \end{equation}
        Therefore, we have $\|\tilde{W}_{\overline{i},:}\|_2<\epsilon$ and thus $W\in B(X_{tgt},\epsilon)$.

    \end{proof}

\section{Proof of \cref{thm:projection}}
\ProjThm*
\begin{proof}[Proof of \cref{thm:projection}]\label{appendix:ProjThm}
    Recall that 
    \begin{equation}
        prune(P)(W,b_W,A,b_A,D,\overline{D}) = (W[P^c],b_W[P^c],(A^T[P^c])^T, b_A',0,\overline{D}[P^c])
    \end{equation}
    where $b_A'=b_A+ADb_W+(A^T[P])^T(\psi_{-\overline{D},0}(b_W))[P]$.

    Let $|P|=n$ and WLOG let $P=\{1,\cdots,n\}$. Then we can write each parameters can be written by block matrix representations:

    \begin{equation}\label{eqn:block-repr}
        A=\left[\begin{array}{c|c}
             A^T[P]^T&A^T[P^c]^T 
        \end{array}\right], W=\left[\begin{array}{c}
             0\\  W[P^c]
        \end{array}\right], b_W = \left[\begin{array}{c}
             b_W[P]\\  b_W[P^c]
        \end{array}\right]
    \end{equation}

    Let $X_{tgt}$ be arbitrary input tensor of $\overline{\varphi_2}(\overline{\theta},D,\overline{D})$ then the output would be given by 
    \begin{equation}
    \begin{aligned}
        \overline{\varphi_2}(\overline{\theta},D,\overline{D})(x)&=b_A+A\psi_{D,\overline{D}}(b_W+Wx)\\
  &=b_A+ADb_W+A\cancelto{0}{DW}x-A\overline{D}b_W-A\overline{D}Wx+A\sigma(b_W+Wx)\\
  &=b_A+ADb_W+A\psi_{0,\overline{D}}(b_W+Wx)
    \end{aligned}
    \end{equation}

    Since $\psi_{0,\overline{D}}$ is channel-wise operation, $\psi_{-\overline{D},0}(b_W)[P] = \psi_{0,\overline{D}[P]}(b_W[P])$. Considering the block matrix representation \cref{eqn:block-repr}, $A\psi_{0,\overline{D}}(b_W+Wx)$ becomes
    \begin{equation}
    \begin{aligned}
        &A\psi_{0,\overline{D}}(b_W+Wx)\\
        &=\left[\begin{array}{c|c}
             A^T[P]^T&A^T[P^c]^T \end{array}\right]\psi_{0,\overline{D}}\biggl(\left[\begin{array}{c}
             b_W[P]\\  b_W[P^c]+W[P^c]x
        \end{array}\right]\biggr)\\
        &=\left[\begin{array}{c|c}
             A^T[P]^T&A^T[P^c]^T 
        \end{array}\right]\left[\begin{array}{c}
             \psi_{0,\overline{D}[P]}(b_W[P])\\  \psi_{0,\overline{D}[P^c]}(b_W[P^c]+W[P^c]x)
        \end{array}\right]\\
        &= A^T[P]^T\psi_{0,\overline{D}[P]}(b_W[P]) \\
        &\hspace{1cm}+ A^T[P^c]^T\psi_{0,\overline{D}[P^c]}(b_W[P^c]+W[P^c]x)
    \end{aligned}
    \end{equation}

Hence, letting $b_A'=b_A+ADb_W+(A^T[P])^T(\psi_{-\overline{D},0}(b_W))[P]$ we finish the proof by
\begin{equation}
    \begin{aligned}
        \overline{\varphi_2}(\overline{\theta},D,\overline{D})(x) &= b_A'+A^T[P^c]^T\psi_{0,\overline{D}[P^c]}(b_W[P^c]+W[P^c]x)\\
  &=\overline{\varphi_2}(b_W[P^c],W[P^c],A^T[P^c]^T,b_A',0,\overline{D}[P^c])\\
  &=(\overline{\varphi_2}\circ prune)(\overline{\theta},D,\overline{D})(x)
    \end{aligned}
\end{equation}
\end{proof}

\section{Proof of \cref{thm:DW-dynamics}}\label{appendix:DWdynamicsThm}
\DWdynamicsThm*

\begin{proof}[Proof of \cref{thm:DW-dynamics}]
    First consider the gradient descent movement of $d_t$ and $M_t^{(i)}$, the $i$th entry of $M_t$, as follows:
    \begin{equation}
        \begin{aligned}\label{eqn:DW-GD}
            d_{t+1} &= d_t - \alpha d_t -sgn(d_t)\lambda_t\|M_t\|_2 = (1-\alpha-\frac{\lambda_t}{c_t})d_t\\
            M_{t+1}^{(i)} &= M_t^{(i)} - \alpha M_t^{(i)}-\lambda_t|d_t|\cdot \frac{M_t^{(i)}}{\|M_t\|_2} = (1-\alpha-\lambda_tc_t)M_t^{(i)}.
        \end{aligned}
    \end{equation}
    Note that the second inequalities of each are induced from the definition of $c_t = \frac{|d_t|}{\|M_t\|_2}$. 

    From second equation of \cref{eqn:DW-GD} we can induce following vector-formed updates:
    \begin{equation}
        M_{t+1} = (1-\alpha-\lambda_tc_t)M_t.
    \end{equation}

    Therefore, if 
    \begin{equation}\label{eqn:lambda_t_cond_ct}
        \frac{\lambda_t}{1-\alpha}\leq c_t\leq \frac{(1-\alpha)}{\lambda_t}
    \end{equation}
    then we have

    \begin{equation}\label{eqn:DW-GD-abs1}
        \|M_{t+1}\|_2 = (1-\alpha-\lambda_tc_t)\|M_t\|_2.
    \end{equation}
    and
    \begin{equation}\label{eqn:DW-GD-abs2}
        d_{t+1} = (1-\alpha-\frac{\lambda_t}{c_t})d_t.
    \end{equation}

    Therefore, we get
    \begin{equation}\label{eqn:ct_induction}
        c_{t+1} = \frac{1-\alpha-\frac{\lambda_t}{c_t}}{1-\alpha-\lambda_tc_t}c_t = f(c_t,\lambda_t)c_t
    \end{equation}
    where $f(x,y) = \frac{1-\alpha-\frac{y}{x}}{1-\alpha-xy}$.

    \begin{enumerate}
        \item[(1)] Suppose $c_0=1$ then simple induction shows that $c_t=1$ for all time $t$ since \cref{eqn:lambda_t_cond_ct} holds by assumption in \cref{eqn:lambda_t_cond_c0}. 

        \item[(2)] Suppose $c_0<1$ and let $T$ be the smallest integer which satisfies $\frac{\lambda_T}{1-\alpha}>c_T$.

        If $t<T$ and $c_t<1$, then $f(c_t,\lambda_t)<1$ by \cref{lem:f_N}(1) and thus $c_{t+1}<c_t<1$, which means that $\{c_t\}$ is a decreasing sequence. 

        Also, by \cref{lem:f_N}(1) we have $f(c_t,\lambda_t)<f(c_0,\lambda_*)<1$ since $c_t<c_0$ because $\{c_t\}$ is a decreasing sequence, and $\lambda_t>\lambda_*$ due to the assumption in \cref{eqn:lambda_t_cond_c0}. Therefore, we have
        \begin{equation}
            c_t< f(c_0,\lambda_*)c_{t-1}<\cdots < f(c_0,\lambda_*)^tc_0
        \end{equation}
        which finishes the proof of (2).

        \item[(3)] Suppose $c_0>1$ and let $T$ be the smallest integer which satisfies $\frac{\lambda_T}{1-\alpha}<c_T$. 

        If $t<T$ and $c_t>1$, then $f(c_t,\lambda_t)>1$ by \cref{lem:f_N}(2) and thus $c_{t+1}>c_t>1$, which means that $\{c_t\}$ is a increasing sequence.

        Also, by \cref{lem:f_N}(2) we have $f(c_t,\lambda_t)>f(c_0,\lambda_*)>1$ since $c_t>c_0$ because $\{c_t\}$ is a increasing sequence, and $\lambda_t>\lambda_*$ due to the assumption in \cref{eqn:lambda_t_cond_c0}. Therefore, we have
        \begin{equation}
            c_t>f(c_0,\lambda_*)c_{t-1}>\cdots > f(c_0,\lambda_*)^tc_0
        \end{equation}
        which completes the proof of (3).

    \end{enumerate}
    
\end{proof}

\begin{lemma}\label{lem:f_N}
    Let $f(x,y) = \frac{1-\alpha-\frac{y}{x}}{1-\alpha-xy}$ and $y\in(0,1-\alpha)$.
    \begin{enumerate}
        \item[(1)] If $x\in (0,1)$ then $f(x,y) < 1$, $\frac{\partial f}{\partial x} > 0$ and $\frac{\partial f}{\partial y} < 0$
        \item[(2)] If $x\in (1,\infty)$ then $f(x,y) > 1$, $\frac{\partial f}{\partial x} > 0$ and $\frac{\partial f}{\partial y} > 0$
    \end{enumerate}

    \begin{proof}
        We first compute the partial derivatives:
        \begin{equation}\label{eqn:fN_derivative_x}
            \frac{\partial f}{\partial x} = \frac{y(1-\alpha)}{x^2(1-\alpha-xy)^2}\{(x-\frac{y}{1-\alpha})^2-\frac{y^2}{(1-\alpha)^2}+1\}
        \end{equation}
        \begin{equation}
            \frac{\partial f}{\partial y} = \frac{(1-\alpha)(x^2-1)}{x(1-\alpha-1xy)^2}
        \end{equation}

        Since $y\in (0,1-\alpha)$, the $-\frac{y^2}{(1-\alpha)^2}+1$ term in RHS of \cref{eqn:fN_derivative_x} becomes positive. Therefore, $\frac{\partial f}{\partial x}>0$ for all $X_{tgt}$.

        Now assume that $x\in (0,1)$. Then $f(x,y)<1$ since $x<\frac{1}{x}$ and $\frac{\partial f}{\partial y} < 0$ because $x^2-1<0$.

        Similarly, if $x\in (1,\infty)$ then $f(x,y)>1$ and $\frac{\partial f}{\partial y}>0$.
        \end{proof}
\end{lemma}

\section{Details on experiments}\label{appendix:hyperparameters}
In this section, we list the details on expeirment designs.

For Resnet-56 and VGG-19, we utilize the implementation and pre-trained models from \cite{fang2023depgraph}, which achieve 93.53\% and 73.50\% top-1 accuracy, respectively. 
For Resnet-50, we use official torchvision \cite{torchvision2016} base model with pre-trained weights, which has 76.15\% top-1 accuracy. 
Standard data augmentation including random cropping and flipping were applied. %, which is same to the pre-training stage. 

All experiments in this paper were conducted with Linux (Ubuntu 20.04) computer equipped with single RTX 4090 GPU with 24GB VRAM. For imagenet experiment, we use gradient accumulation to run with single GPU.
\begin{table}[H]
\begin{tabular}{llll}
\toprule
& Resnet56+CIFAR10 & VGG19+CIFAR100& Resnet50+Imagenet \\
\midrule
batch size& 128 & 128 & 64 \\
Gradient accumulation& 1& 1& 2  \\
optimizer & SGD(momentum=0.9)& SGD(momentum=0.9)& SGD(momentum=0.9) \\
Weight decay ($\alpha_\theta,\alpha_D$) & (5e-4,5e-5)& (5e-4,0)  & (1e-4,0)\\
c& 1& 1& 1  \\
$\gamma_t$& 0.007(1+0.25t),  & $\gamma$(1+0.25t), $\gamma$=[2e-3,9e-3,12e-3] & 3e-4(1+0.25t)   \\
$\epsilon$ (opt1,opt2)& 1e-6,1e-6 & 2e-6,1e-6 & 3e-6,3e-6    \\
Stage finish epochs & \multirow{2}{*}{100,200,300}  & \multirow{2}{*}{200,300,400}  & \multirow{2}{*}{20,40,170}   \\
(opt1,opt2,finetune)& & &    \\
LR  & \multirow{2}{*}{1e-2,1e-2} & 5e-3,5e-3, 1e-3 for 3x  & \multirow{2}{*}{5e-3,5e-3,(5e-3,1e-5)}  \\
(opt1,opt2,finetune)& & 0.01, 0.01, 0.01  &    \\
LR decay epoch& \multirow{2}{*}{[50],[150],[240,270]} & [],[],[390] for 3x& \multirow{2}{*}{[10,15],[30,35],[70,100,120,160]} \\
(opt1,opt2,finetune)& & [75,750],[250],[340,370] for rest  &    \\
LR decay ratio& \multirow{2}{*}{0.1, 0.1, 0.1}& NA,NA,0.1 for 3x & \multirow{2}{*}{0.1,0.1,0.1}\\
(opt1,opt2,finetune)& & 0.2,0.2,0.1 for rest&    \\
training runtime & 2 hours & 2 hours &60.42hours \\
\bottomrule                                           
\end{tabular}
\end{table}
\newpage
\section{Loss curves during the training} \label{sec:loss_curves_during}
In this section, we plot the training logs of loss, accuracy, $\|DW\|$ and MACs, in CIFAR10 experiment. The target layers of the model were pruned early in epoch 51 and 116.

\begin{figure}[H]
  \centering
    \begin{subfigure}[b]{0.99\textwidth}
    \centering
      \includegraphics[width=\textwidth]{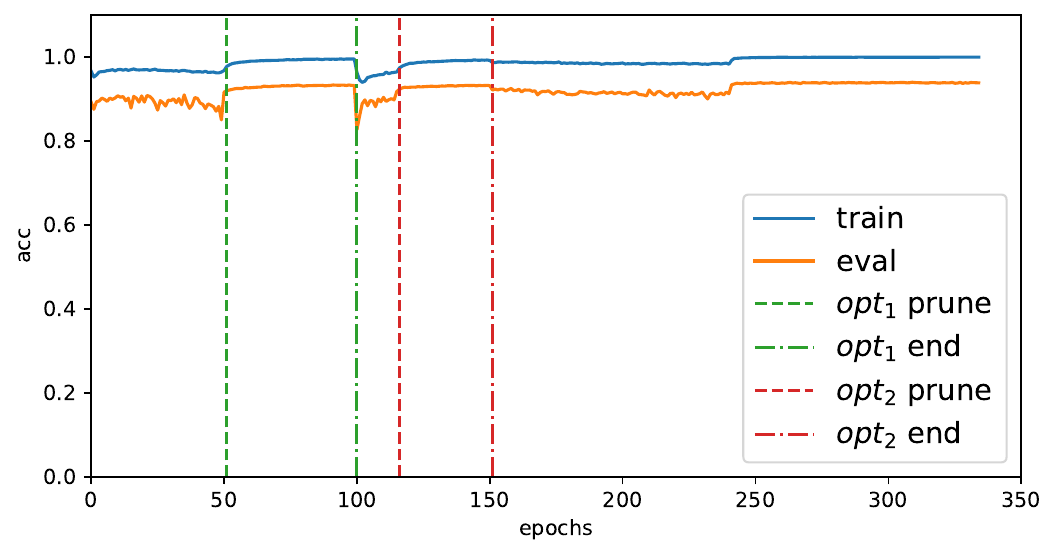}
      \caption{Train and Test Accuracy}
      \label{subfig:learning_curve_acc}
    \end{subfigure}%
    \hfill%
    \begin{subfigure}[b]{0.49\textwidth}
    \centering
      \includegraphics[width=\textwidth]{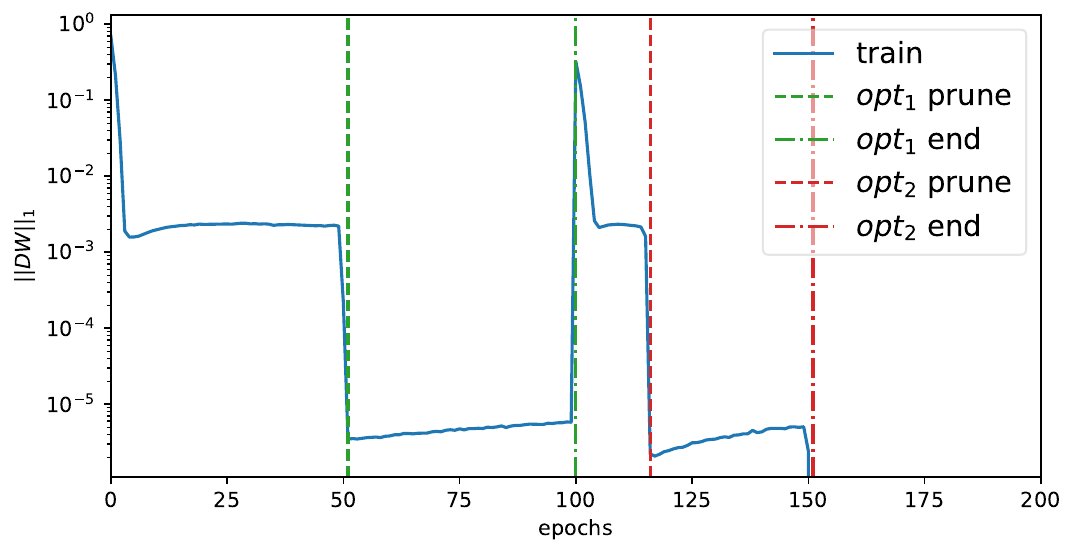}
      \caption{total $\|DW\|_{2,1}$}
      \label{subfig:learning_curve_DW}
    \end{subfigure}%
    \hfill%
    \begin{subfigure}[b]{0.49\textwidth}
    \centering
      \includegraphics[width=\textwidth]{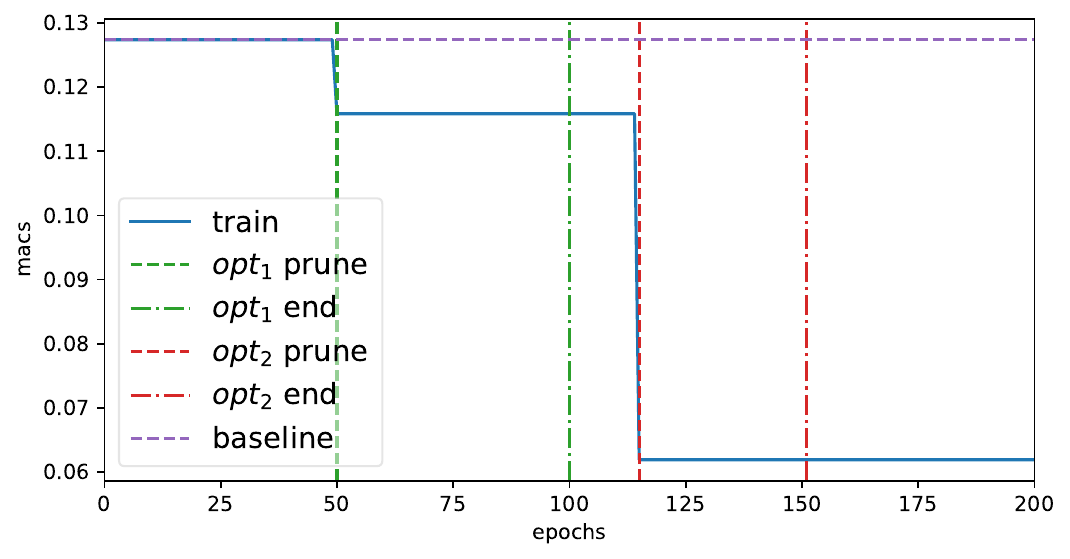}
      \caption{MACs reduction}
      \label{subfig:learning_curve_macs}
    \end{subfigure}%
    \hfill%
    \begin{subfigure}[b]{0.49\textwidth}
    \centering
      \includegraphics[width=\textwidth]{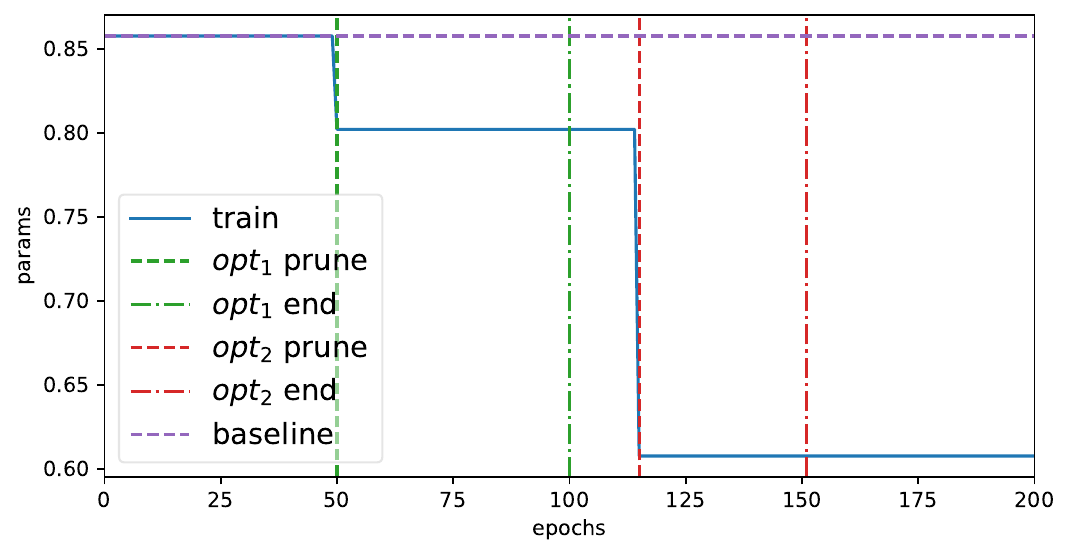}
      \caption{number of parameters}
      \label{subfig:learning_curve_params}
      
    \end{subfigure}%
    \hfill%
    \begin{subfigure}[b]{0.49\textwidth}
    \centering
      \includegraphics[width=\textwidth]{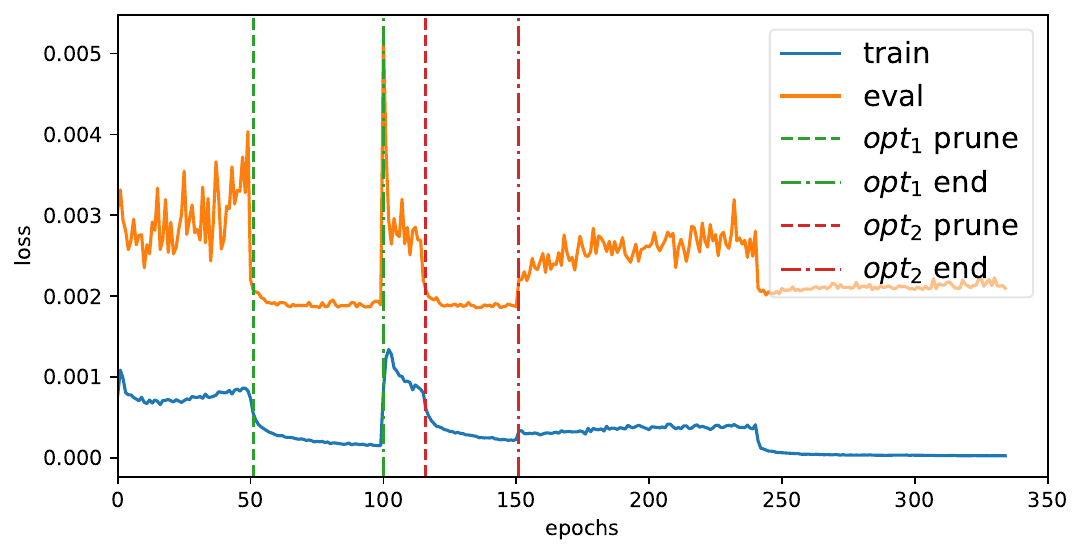}
      \caption{Loss $\mathcal{L}_2$}
      \label{subfig:learning_curve_loss}
    \end{subfigure}%
\hfill%
\caption{Learning curves of CIFAR10 experiment with Resnet56}

\label{fig:loss_curve}
\end{figure}

\clearpage

\section{Full evaluation on bifurcation behavior}\label{appendix:plots_bifurcation}
In this section, we provide histograms of filter norm, $\{D_{ii}\}_{i\in [N_W]}$ and $C = \frac{D}{\|W\|_{2,1}}$ for every layer of our pruned VGG19 model (in \cref{subsec:res-performance}) and Resnet50 model(speedup 2.00x), to show that the bifurcation behavior claimed in \cref{subsec:bifurcation} always happens.

\subsection{Resnet56+CIFAR10}
\begin{figure*}[h]
  \centering
    \begin{subfigure}[b]{0.33\textwidth}
    \centering
      \includegraphics[width=\textwidth]{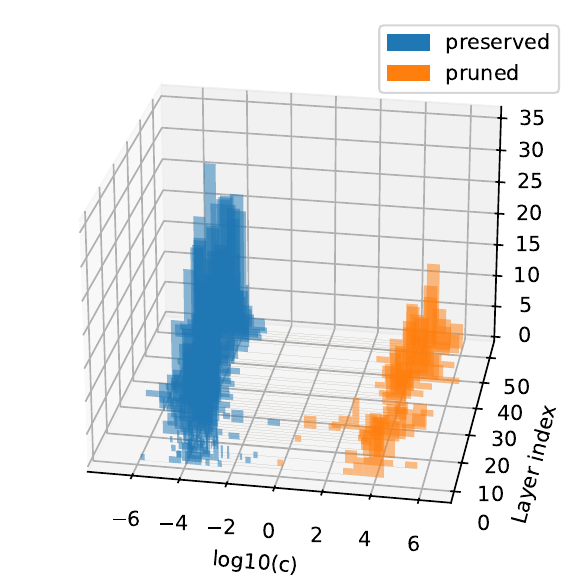}
      \caption{Bifurcation on $c_i$'s}
      \label{subfig:r56_polar_C}
    \end{subfigure}%
    \hfill%
    \begin{subfigure}[b]{0.33\textwidth}
    \centering
      \includegraphics[width=\textwidth]{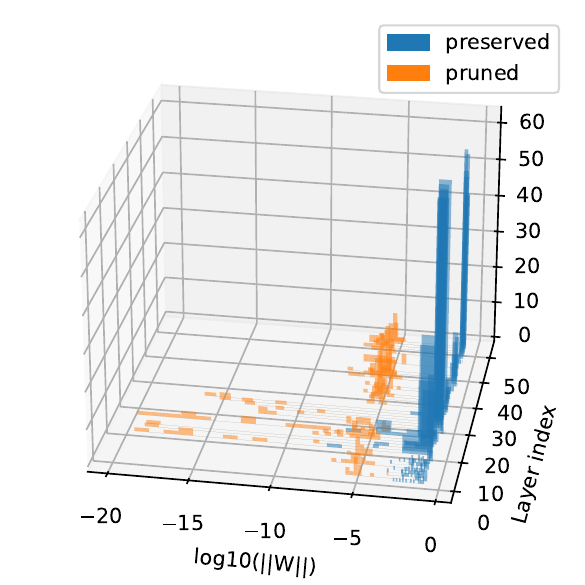}
      \caption{Bifurcation on $F_i$s}
      \label{subfig:r56_polar_W}
    \end{subfigure}%
    \hfill%
    \begin{subfigure}[b]{0.33\textwidth}
    \centering
      \includegraphics[width=\textwidth]{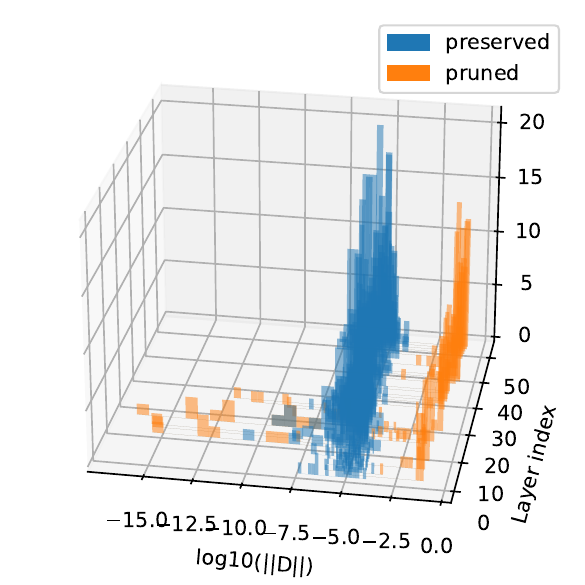}
      \caption{Bifurcation on $D_{ii}$s}
      \label{subfig:r56_polar_D}
    \end{subfigure}%
\hfill%
\caption{The histograms of ratio $c=\frac{D_{ii}}{\|F_i\|}$, filter vector $F_i$ of weight tensor $W$, and $D_{ii}$'s for each layers of Resnet56 model trained on CIFAR10. The z-axis represents the frequency. The layer indices of results in $opt2$ are shifted to start from the last layer.}.
\label{fig:r56x2_polarization}
\end{figure*}

\subsection{VGG19+CIFAR100}
\begin{figure*}[h]
  \centering
    \begin{subfigure}[b]{0.33\textwidth}
    \centering
      \includegraphics[width=\textwidth]{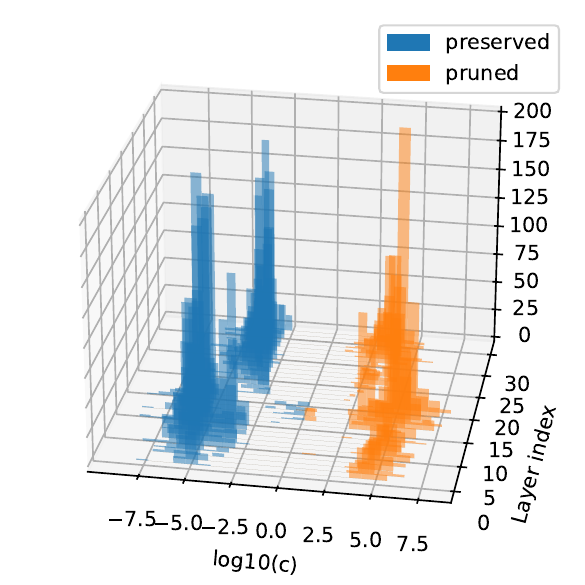}
      \caption{Bifurcation on $c_i$'s}
      \label{subfig:v19x3_polar_C}
    \end{subfigure}%
    \hfill%
    \begin{subfigure}[b]{0.33\textwidth}
    \centering
      \includegraphics[width=\textwidth]{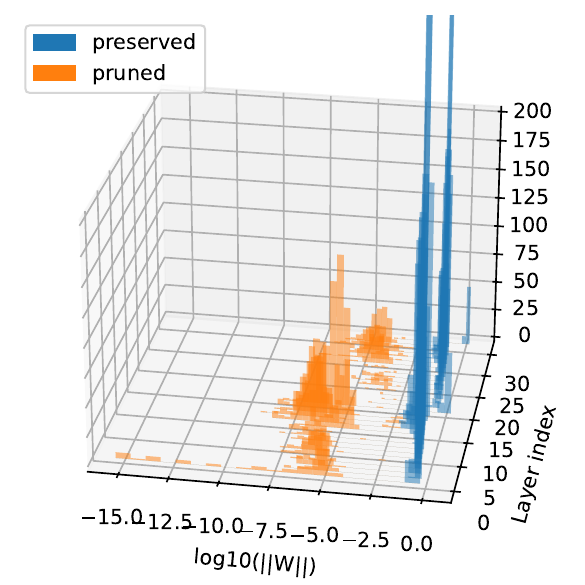}
      \caption{Bifurcation on $F_i$s}
      \label{subfig:v19x3_polar_W}
    \end{subfigure}%
    \hfill%
    \begin{subfigure}[b]{0.33\textwidth}
    \centering
      \includegraphics[width=\textwidth]{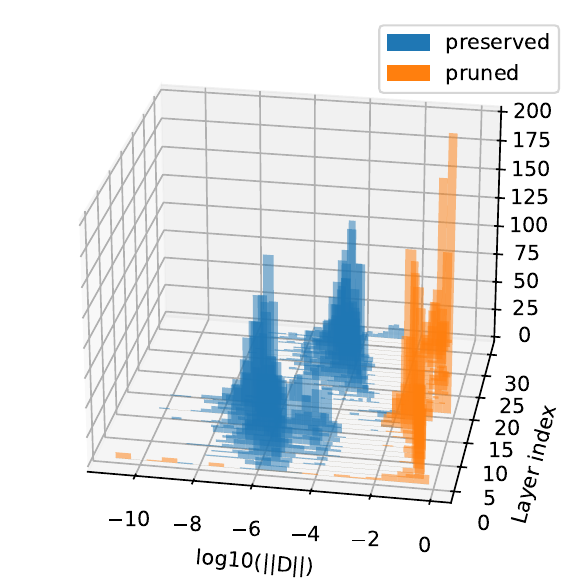}
      \caption{Bifurcation on $D_{ii}$s}
      \label{subfig:v19x3_polar_D}
    \end{subfigure}%
\hfill%
\caption{The histograms of ratio $c=\frac{D_{ii}}{\|F_i\|}$, filter vector $F_i$ of weight tensor $W$, and $D_{ii}$'s for each layers of VGG19 model (3x speedup) trained on CIFAR100. The z-axis represents the frequency. The layer indices of results in $opt2$ are shifted to start from the last layer.}.
\label{fig:v19x3_polarization}
\end{figure*}

\begin{figure*}[h]
  \centering
    \begin{subfigure}[b]{0.33\textwidth}
    \centering
      \includegraphics[width=\textwidth]{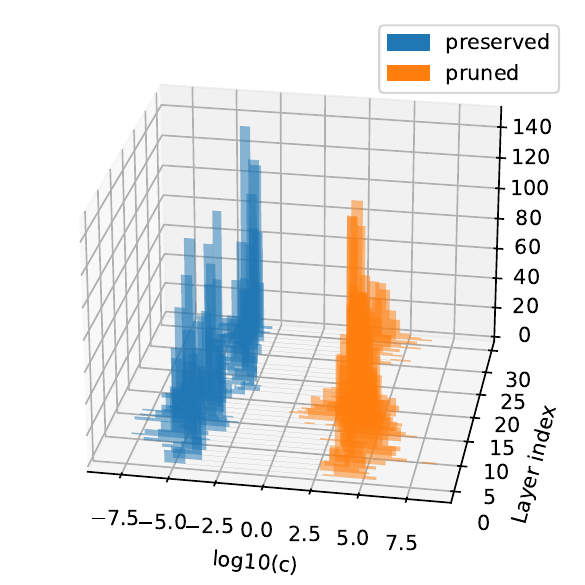}
      \caption{Bifurcation on $c_i$'s}
      \label{subfig:v19x9_polar_C}
    \end{subfigure}%
    \hfill%
    \begin{subfigure}[b]{0.33\textwidth}
    \centering
      \includegraphics[width=\textwidth]{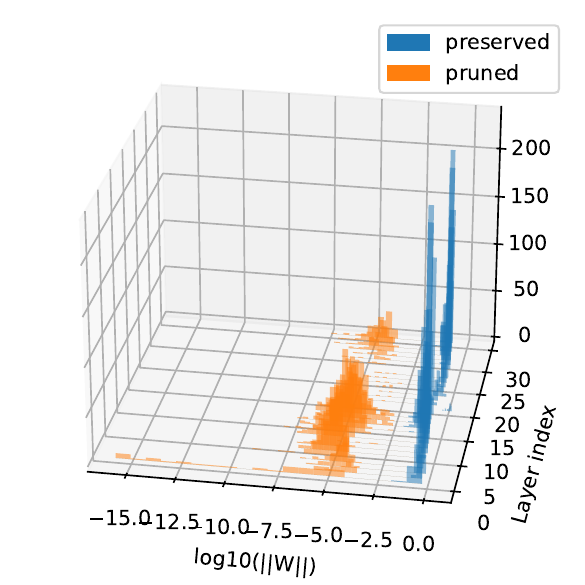}
      \caption{Bifurcation on $F_i$s}
      \label{subfig:v19x9_polar_W}
    \end{subfigure}%
    \hfill%
    \begin{subfigure}[b]{0.33\textwidth}
    \centering
      \includegraphics[width=\textwidth]{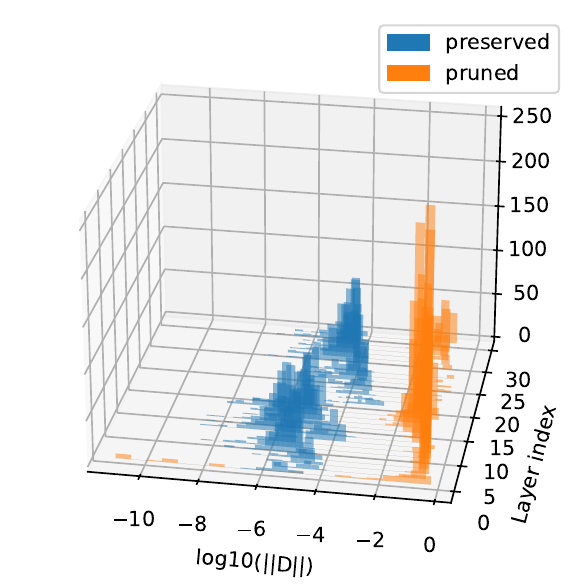}
      \caption{Bifurcation on $D_{ii}$s}
      \label{subfig:v19x9_polar_D}
    \end{subfigure}%
\hfill%
\caption{The histograms of ratio $c=\frac{D_{ii}}{\|F_i\|}$, filter vector $F_i$ of weight tensor $W$, and $D_{ii}$'s for each layers of VGG19 model (8.96x speedup) trained on CIFAR100. The z-axis represents the frequency. The layer indices of results in $opt2$ are shifted to start from the last layer.}.
\label{fig:v19x9_polarization}
\end{figure*}

\begin{figure*}[h]
  \centering
    \begin{subfigure}[b]{0.33\textwidth}
    \centering
      \includegraphics[width=\textwidth]{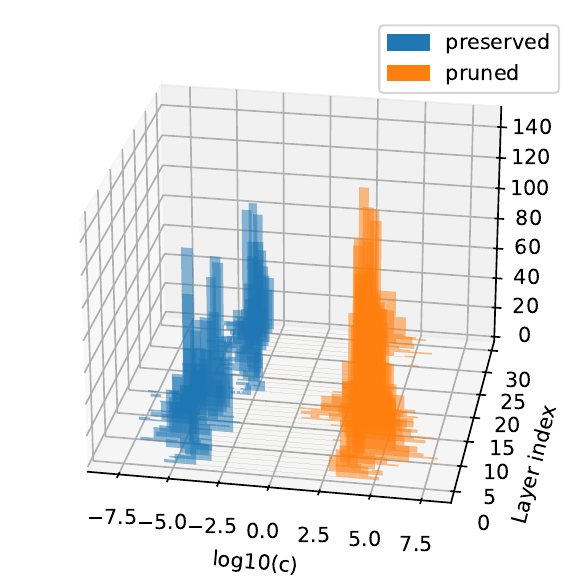}
      \caption{Bifurcation on $c_i$'s}
      \label{subfig:v19x12_polar_C}
    \end{subfigure}%
    \hfill%
    \begin{subfigure}[b]{0.33\textwidth}
    \centering
      \includegraphics[width=\textwidth]{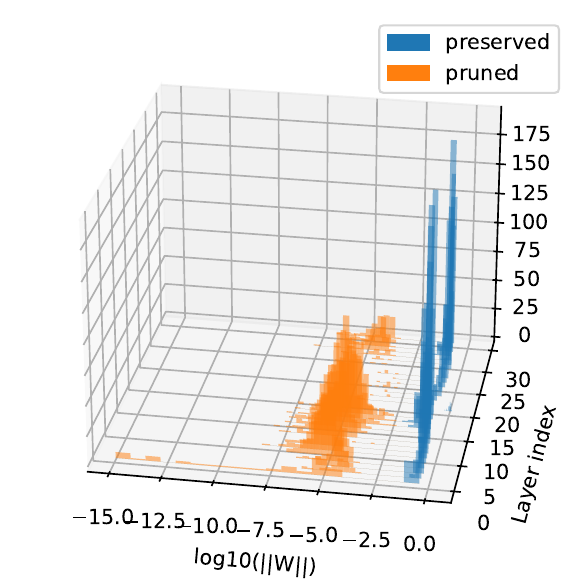}
      \caption{Bifurcation on $F_i$s}
      \label{subfig:v19x12_polar_W}
    \end{subfigure}%
    \hfill%
    \begin{subfigure}[b]{0.33\textwidth}
    \centering
      \includegraphics[width=\textwidth]{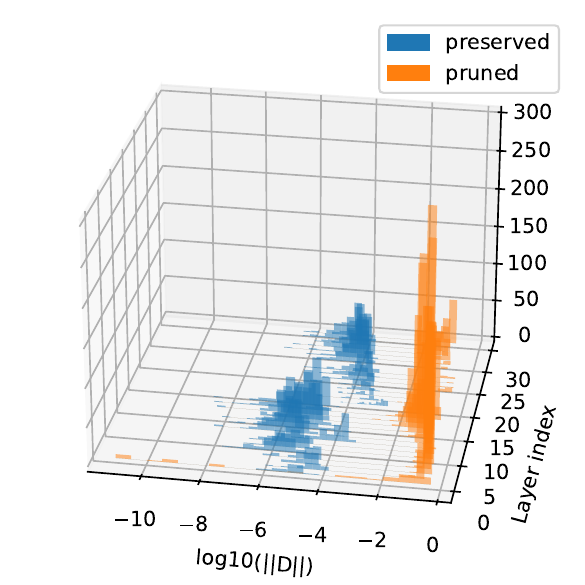}
      \caption{Bifurcation on $D_{ii}$s}
      \label{subfig:v19x12_polar_D}
    \end{subfigure}%
\hfill%
\caption{The histograms of ratio $c=\frac{D_{ii}}{\|F_i\|}$, filter vector $F_i$ of weight tensor $W$, and $D_{ii}$'s for each layers of VGG19 model (11.84x speedup) trained on CIFAR100. The z-axis represents the frequency. The layer indices of results in $opt2$ are shifted to start from the last layer.}.
\label{fig:v19x12_polarization}
\end{figure*}

\newpage
\subsection{Resnet50+Imagenet}
\begin{figure*}[h]
  \centering
    \begin{subfigure}[b]{0.33\textwidth}
    \centering
      \includegraphics[width=\textwidth]{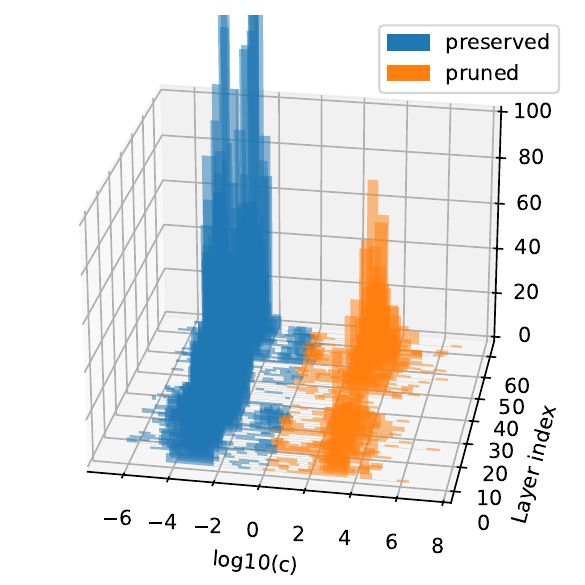}
      \caption{Bifurcation on $c_i$'s}
      \label{subfig:r50x1.49_polar_C}
    \end{subfigure}%
    \hfill%
    \begin{subfigure}[b]{0.33\textwidth}
    \centering
      \includegraphics[width=\textwidth]{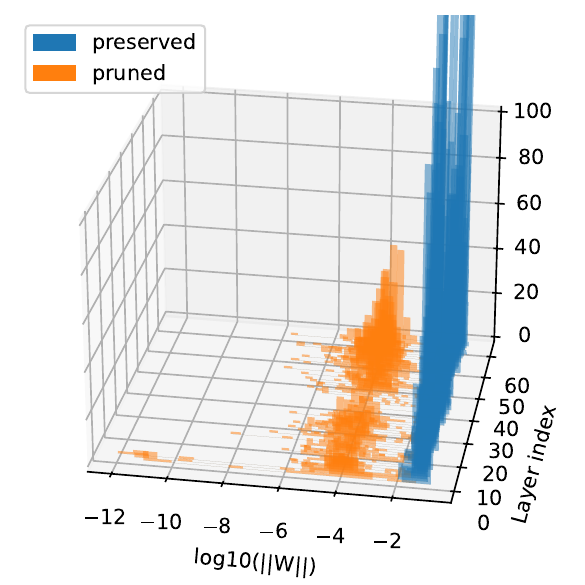}
      \caption{Bifurcation on $F_i$s}
      \label{subfig:r50x1.49_polar_W}
    \end{subfigure}%
    \hfill%
    \begin{subfigure}[b]{0.33\textwidth}
    \centering
      \includegraphics[width=\textwidth]{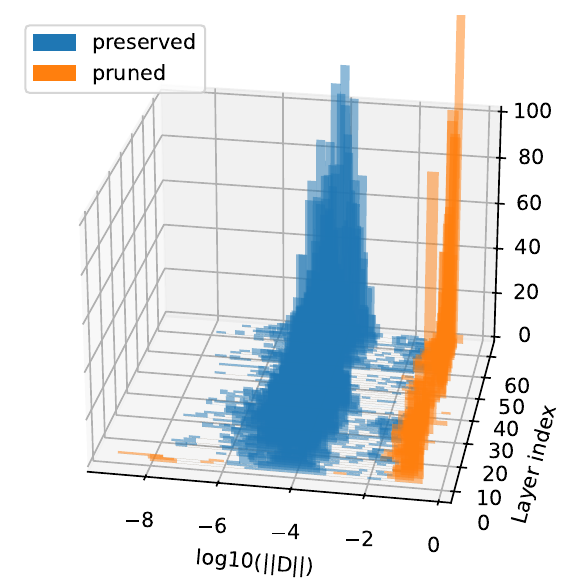}
      \caption{Bifurcation on $D_{ii}$s}
      \label{subfig:r50x1.49_polar_D}
    \end{subfigure}%
\hfill%
\caption{The histograms of ratio $c=\frac{D_{ii}}{\|F_i\|}$, filter vector $F_i$ of weight tensor $W$, and $D_{ii}$'s for each layers of Resnet50 model (1.49x speedup) trained on Imagenet. The z-axis represents the frequency. The layer indices of results in $opt2$ are shifted to start from the last layer.}.
\label{fig:r50x1.49_polarization}
\end{figure*}

\begin{figure*}[h]
  \centering
    \begin{subfigure}[b]{0.33\textwidth}
    \centering
      \includegraphics[width=\textwidth]{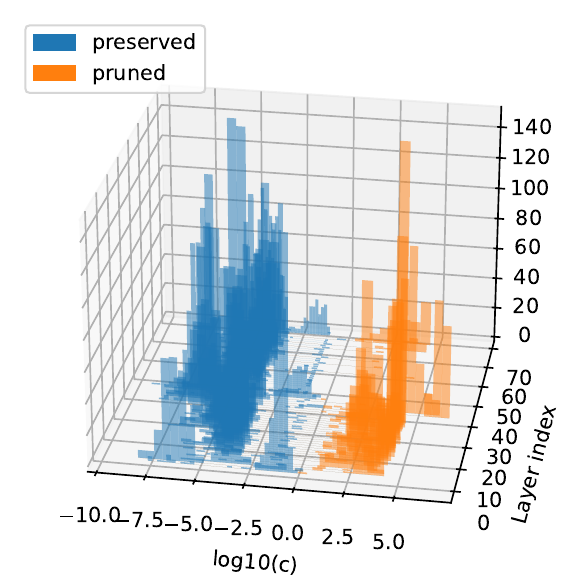}
      \caption{Bifurcation on $c_i$'s}
      \label{subfig:r50x2_polar_C}
    \end{subfigure}%
    \hfill%
    \begin{subfigure}[b]{0.33\textwidth}
    \centering
      \includegraphics[width=\textwidth]{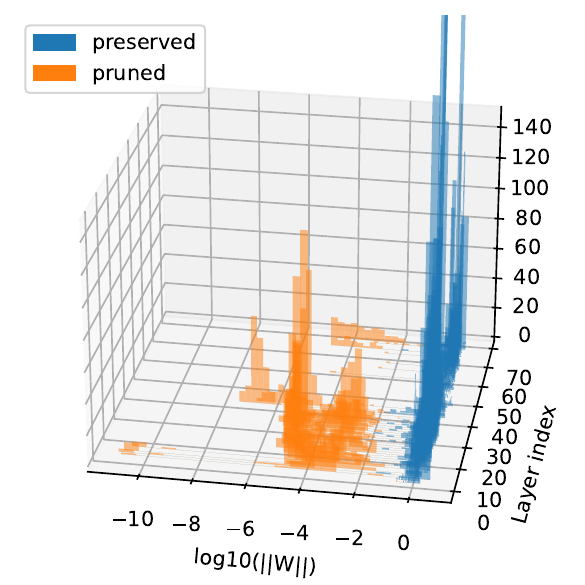}
      \caption{Bifurcation on $F_i$s}
      \label{subfig:r50x2_polar_W}
    \end{subfigure}%
    \hfill%
    \begin{subfigure}[b]{0.33\textwidth}
    \centering
      \includegraphics[width=\textwidth]{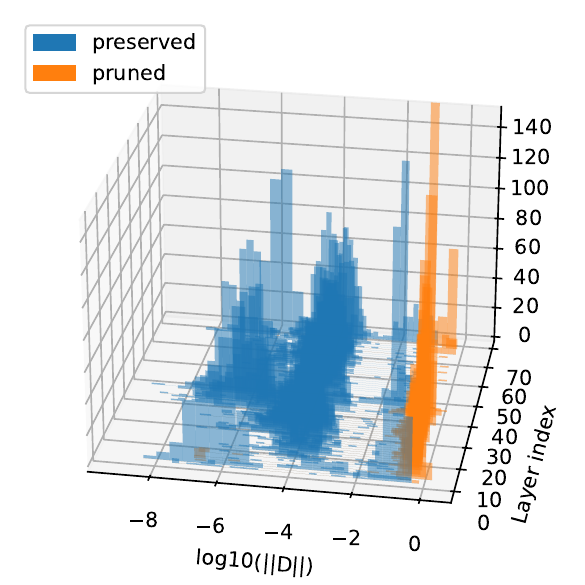}
      \caption{Bifurcation on $D_{ii}$s}
      \label{subfig:r50x2_polar_D}
    \end{subfigure}%
\hfill%
\caption{The histograms of ratio $c=\frac{D_{ii}}{\|F_i\|}$, filter vector $F_i$ of weight tensor $W$, and $D_{ii}$'s for each layers of Resnet50 model (1.96x speedup) trained on Imagenet. The z-axis represents the frequency. The layer indices of results in $opt2$ are shifted to start from the last layer.}.
\label{fig:r50x2_polarization}
\end{figure*}

\begin{figure*}[h]
  \centering
    \begin{subfigure}[b]{0.33\textwidth}
    \centering
      \includegraphics[width=\textwidth]{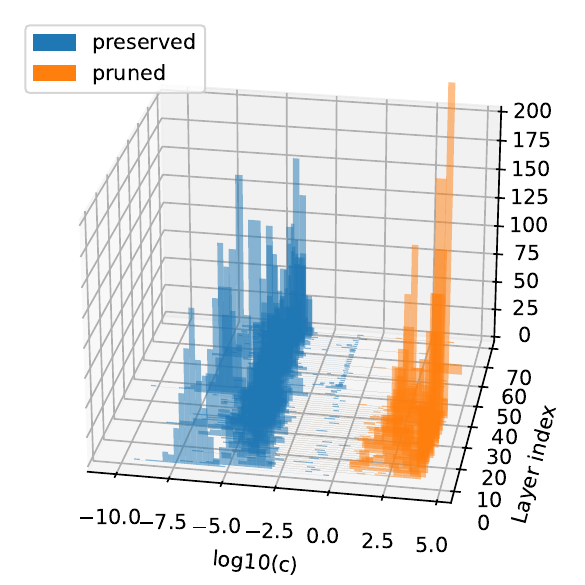}
      \caption{Bifurcation on $c_i$'s}
      \label{subfig:r50x2.3_polar_C}
    \end{subfigure}%
    \hfill%
    \begin{subfigure}[b]{0.33\textwidth}
    \centering
      \includegraphics[width=\textwidth]{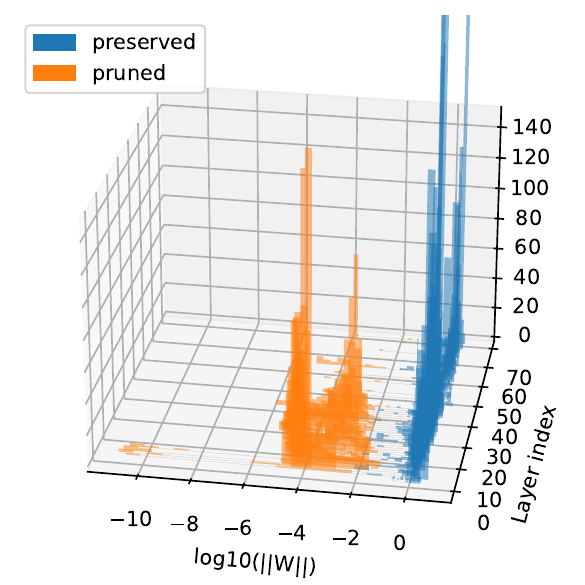}
      \caption{Bifurcation on $F_i$s}
      \label{subfig:r50x2.3_polar_W}
    \end{subfigure}%
    \hfill%
    \begin{subfigure}[b]{0.33\textwidth}
    \centering
      \includegraphics[width=\textwidth]{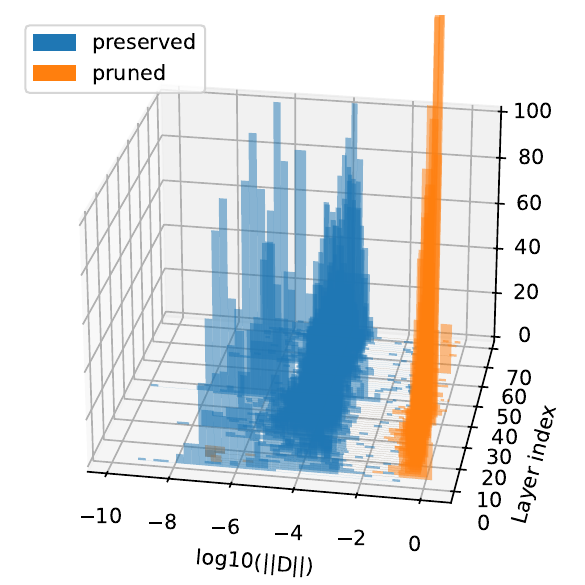}
      \caption{Bifurcation on $D_{ii}$s}
      \label{subfig:r50x2.3_polar_D}
    \end{subfigure}%
\hfill%
\caption{The histograms of ratio $c=\frac{D_{ii}}{\|F_i\|}$, filter vector $F_i$ of weight tensor $W$, and $D_{ii}$'s for each layers of Resnet50 model (2.33x speedup) trained on Imagenet. The z-axis represents the frequency. The layer indices of results in $opt2$ are shifted to start from the last layer.}.
\label{fig:r50x2.33_polarization}
\end{figure*}
\clearpage

\section{Visualizations on fair pruning chance} \label{appendix: plots_init_magnitude_preference}
The L1 and Group Lasso regularizer is claimed to prefer the filters with small initial magnitude. In this section, we regularize VGG19 model on CIFAR100, prune filters according to magnitude with pruning ratio which is same to our pruned VGG19 model (speedup 8.96x in \cref{tab:performance}) and visualize the initial magnitude of pruned filters on every layers.

We also visualize the histogram of filter norms after the regularization too. The bifurcation behaviors can be found on some layers, but still there is preference of small initial magnitude. 

\begin{figure}[H]
    \centering
    \includegraphics[width=\textwidth]{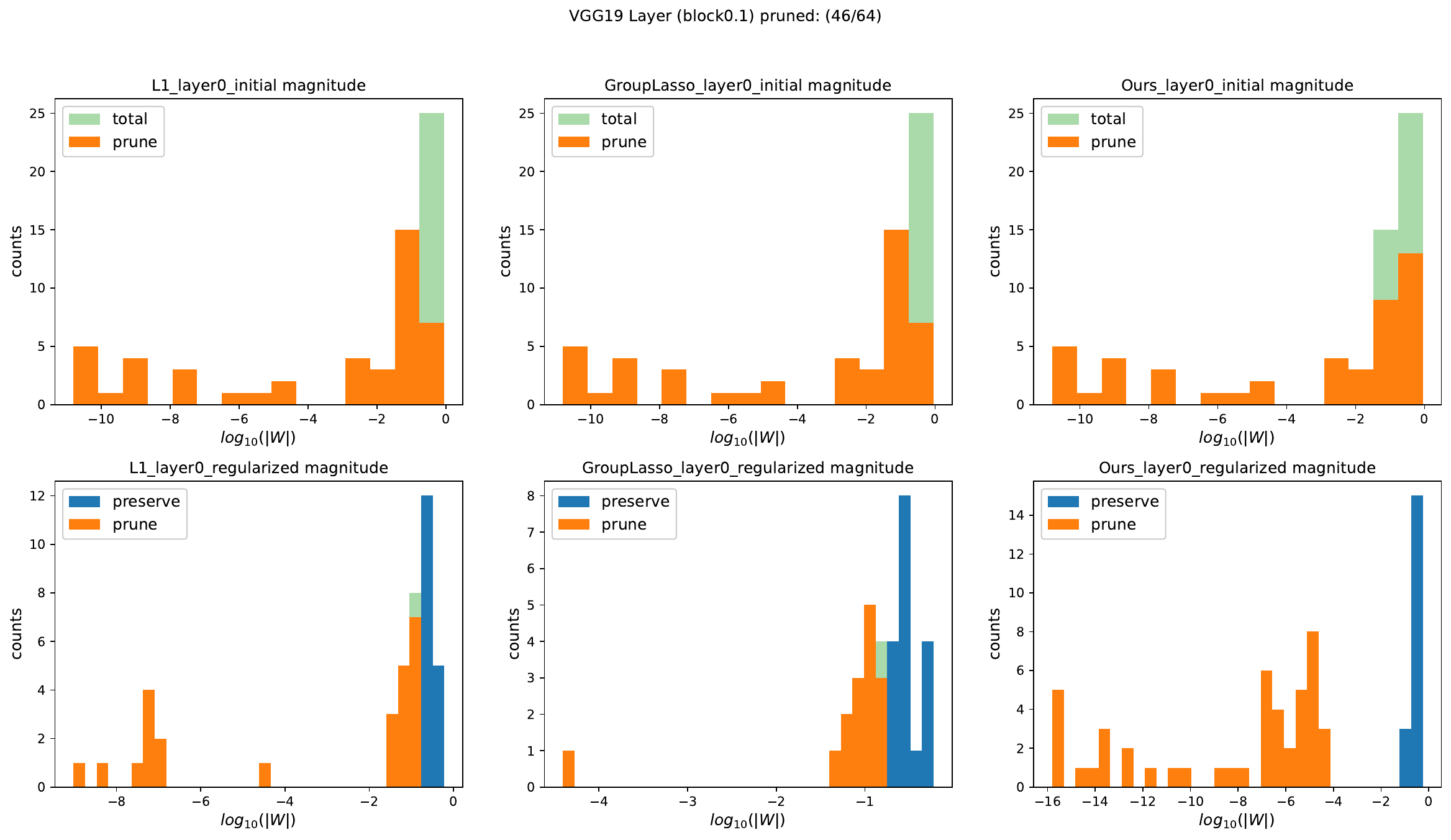}
    \caption{Initial magnitude and regularized magnitude of L1, Group Lass and our regularizer on 1st layer of VGG19 model. Each columns correspond to L1, Group Lasso, $\|DW\|_{2,1}$ respectively. First row is histogram of initial filter norms and second is historgram of regularized filter norms.}
\end{figure}
\begin{figure}[H]
    \centering
    \includegraphics[width=\textwidth]{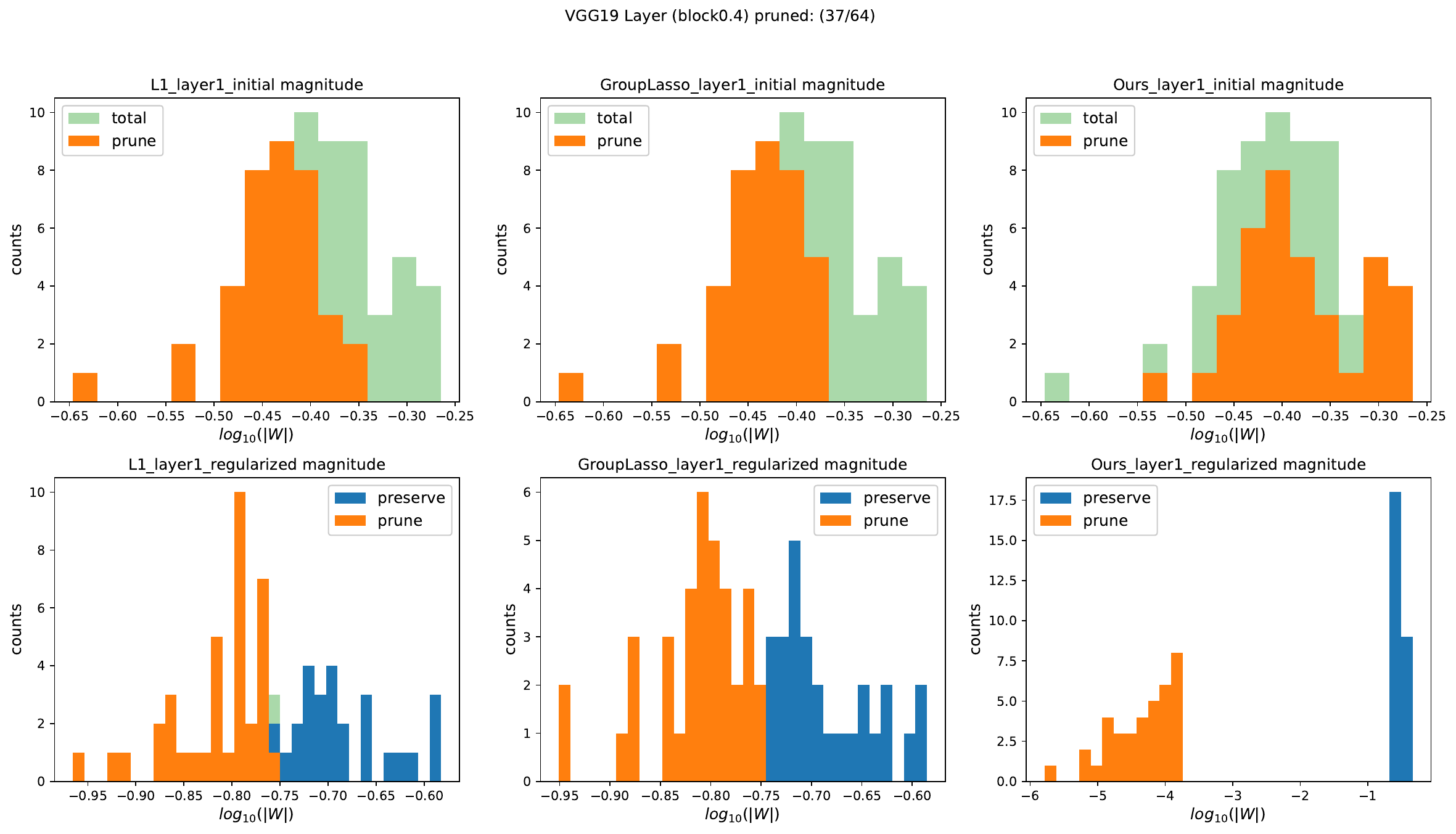}
    \caption{Initial magnitude and regularized magnitude of L1, Group Lass and our regularizer on 2nd layer of VGG19 model. Each columns correspond to L1, Group Lasso, $\|DW\|_{2,1}$ respectively. First row is histogram of initial filter norms and second is historgram of regularized filter norms.}
\end{figure}
\begin{figure}[H]
    \centering
    \includegraphics[width=\textwidth]{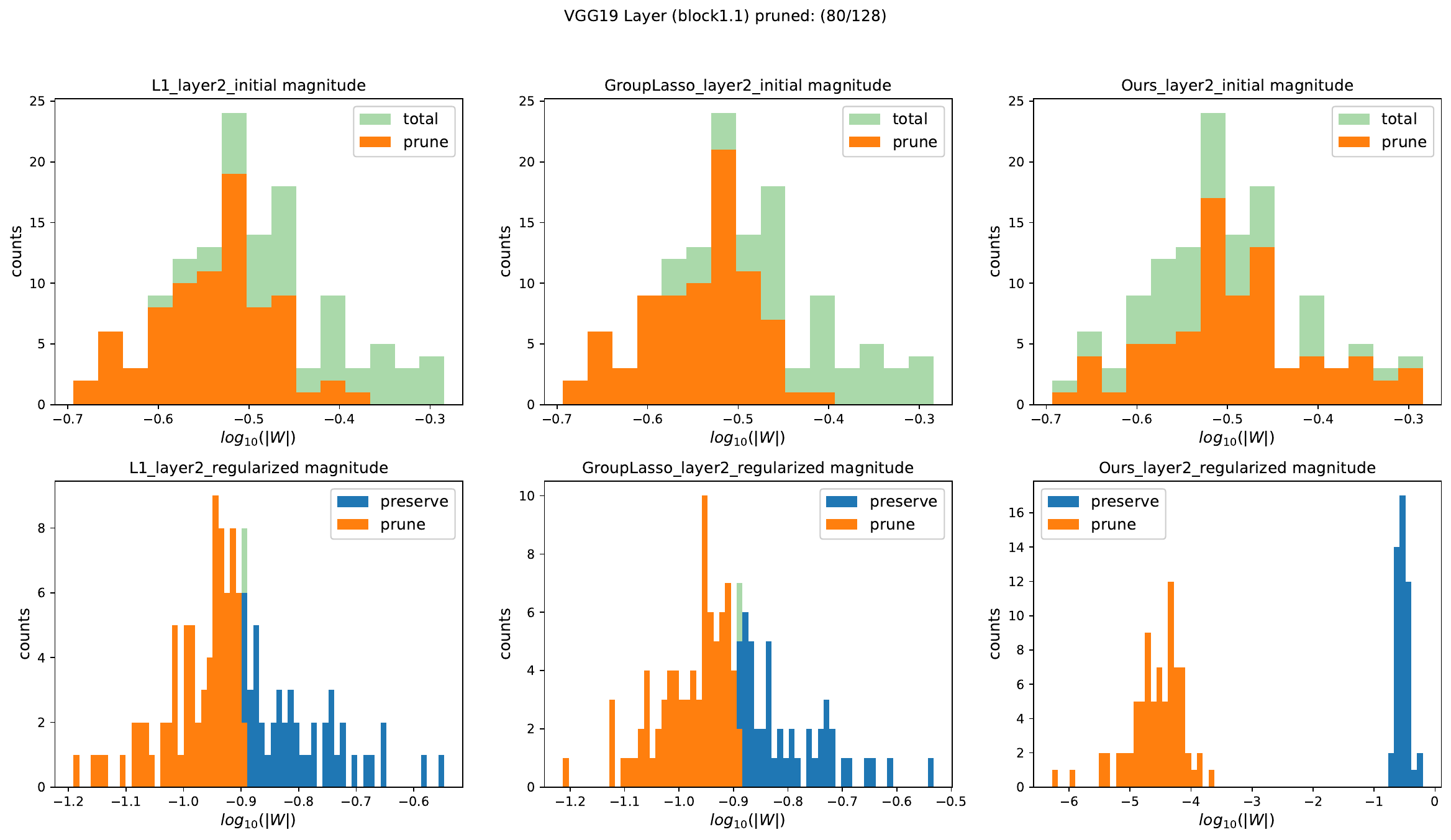}
    \caption{Initial magnitude and regularized magnitude of L1, Group Lass and our regularizer on 3rd layer of VGG19 model. Each columns correspond to L1, Group Lasso, $\|DW\|_{2,1}$ respectively. First row is histogram of initial filter norms and second is historgram of regularized filter norms.}
\end{figure}
\begin{figure}[H]
    \centering
    \includegraphics[width=\textwidth]{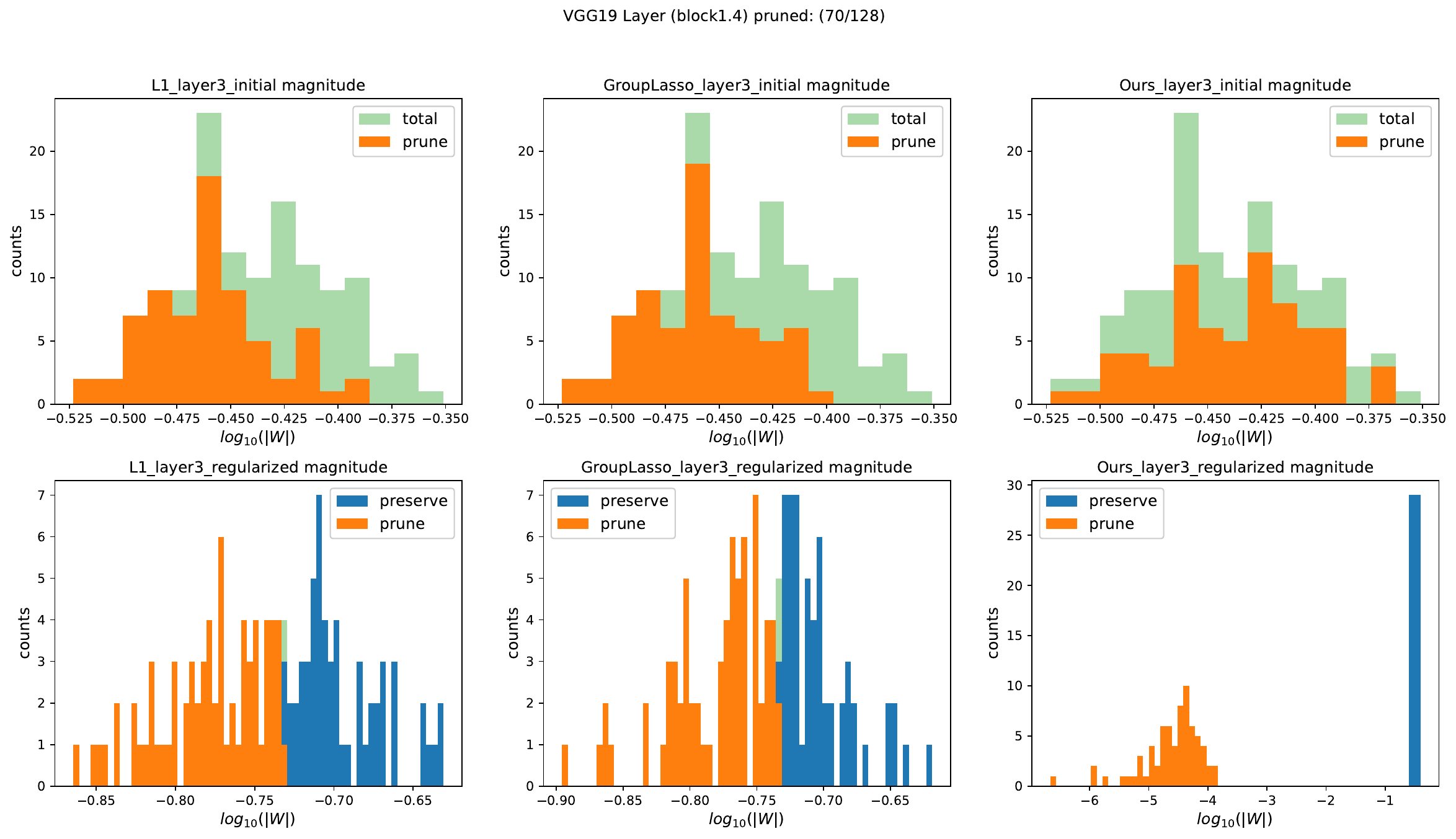}
    \caption{Initial magnitude and regularized magnitude of L1, Group Lass and our regularizer on 4th layer of VGG19 model. Each columns correspond to L1, Group Lasso, $\|DW\|_{2,1}$ respectively. First row is histogram of initial filter norms and second is historgram of regularized filter norms.}
\end{figure}\begin{figure}[H]
    \centering
    \includegraphics[width=\textwidth]{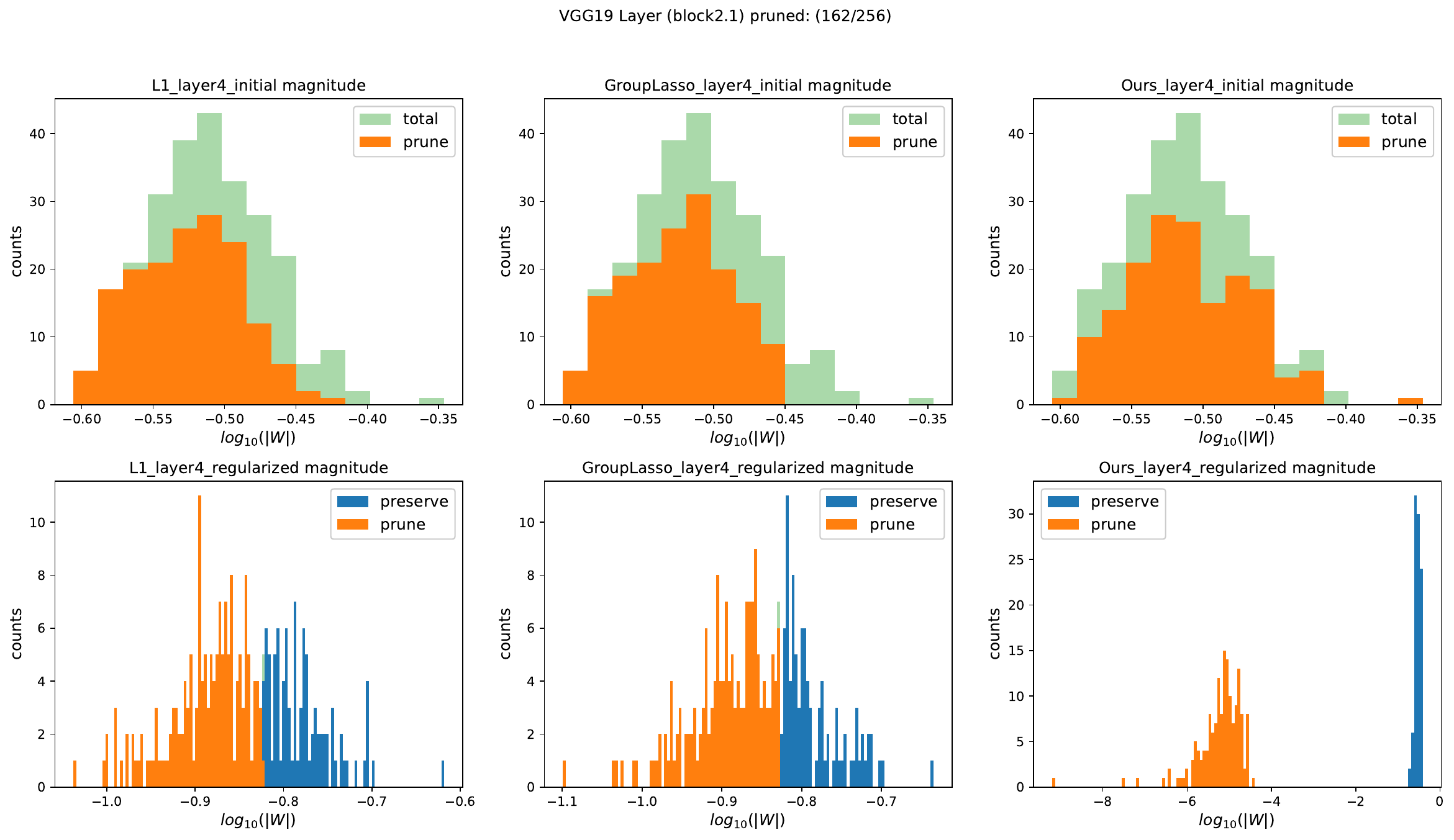}
    \caption{Initial magnitude and regularized magnitude of L1, Group Lass and our regularizer on 5th layer of VGG19 model. Each columns correspond to L1, Group Lasso, $\|DW\|_{2,1}$ respectively. First row is histogram of initial filter norms and second is historgram of regularized filter norms.}
\end{figure}

\begin{figure}[H]
    \centering
    \includegraphics[width=\textwidth]{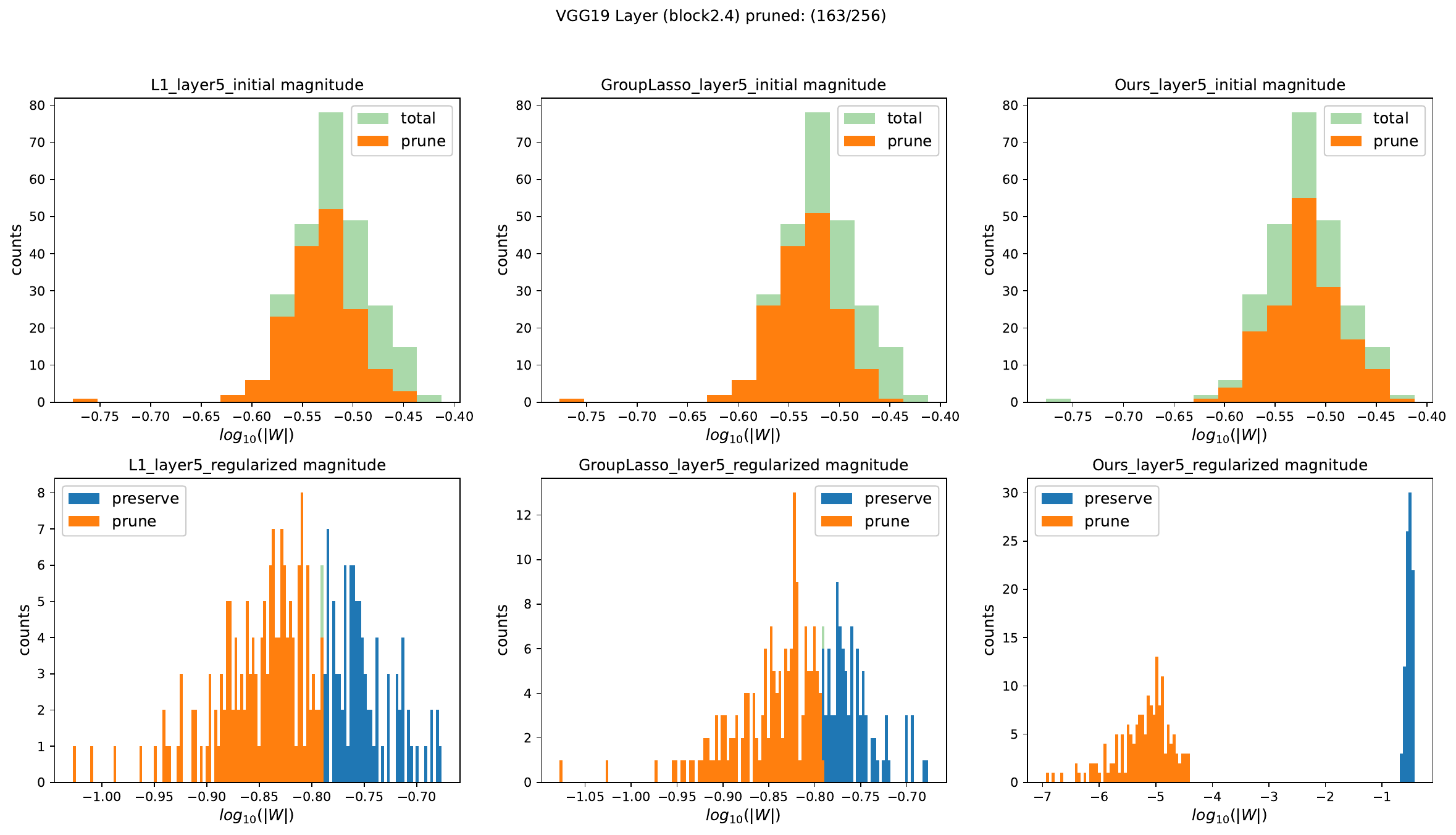}
    \caption{Initial magnitude and regularized magnitude of L1, Group Lass and our regularizer on 6th layer of VGG19 model. Each columns correspond to L1, Group Lasso, $\|DW\|_{2,1}$ respectively. First row is histogram of initial filter norms and second is historgram of regularized filter norms.}
\end{figure}
\begin{figure}[H]
    \centering
    \includegraphics[width=\textwidth]{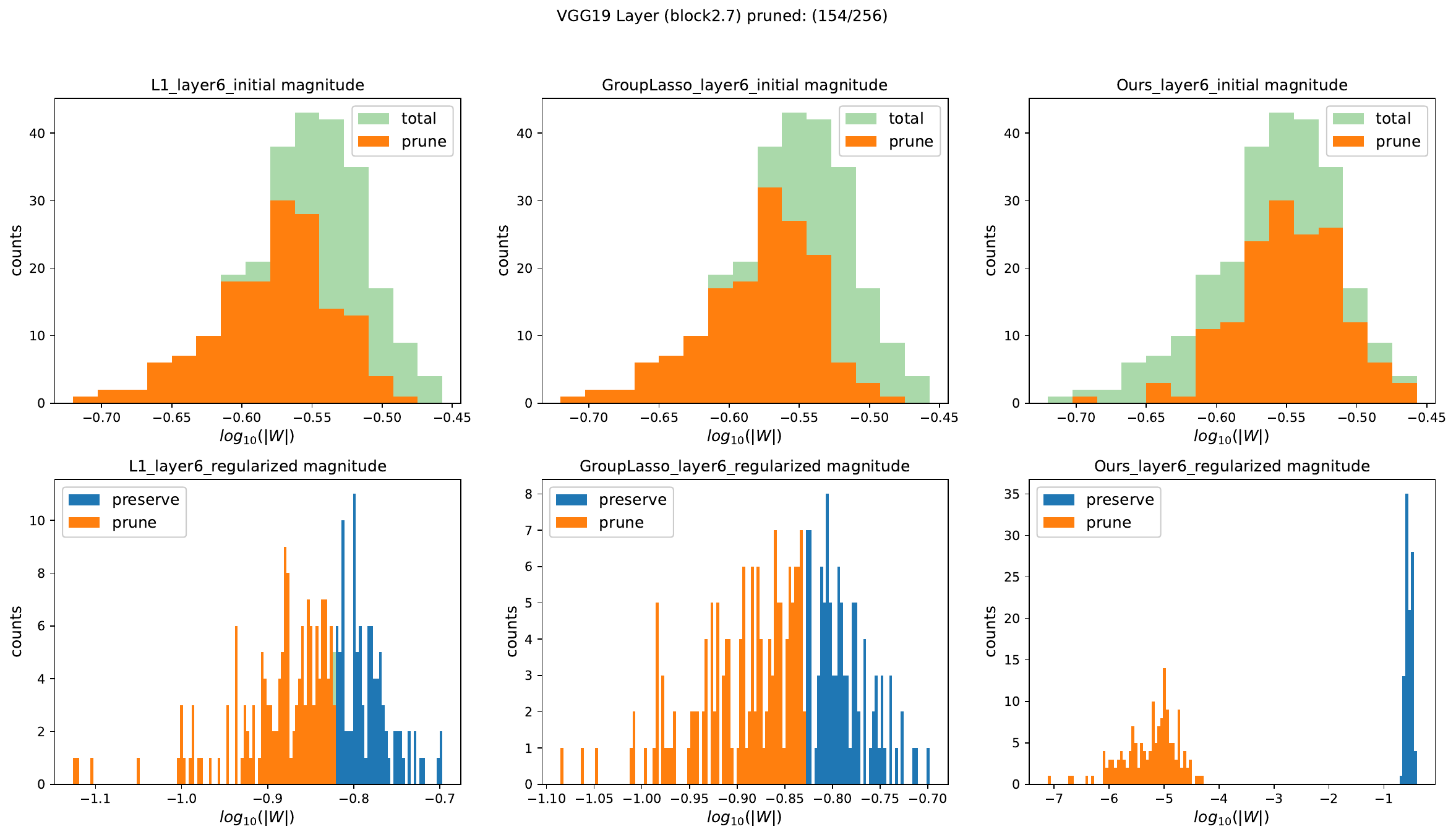}
    \caption{Initial magnitude and regularized magnitude of L1, Group Lass and our regularizer on 7th layer of VGG19 model. Each columns correspond to L1, Group Lasso, $\|DW\|_{2,1}$ respectively. First row is histogram of initial filter norms and second is historgram of regularized filter norms.}
\end{figure}
\begin{figure}[H]
    \centering
    \includegraphics[width=\textwidth]{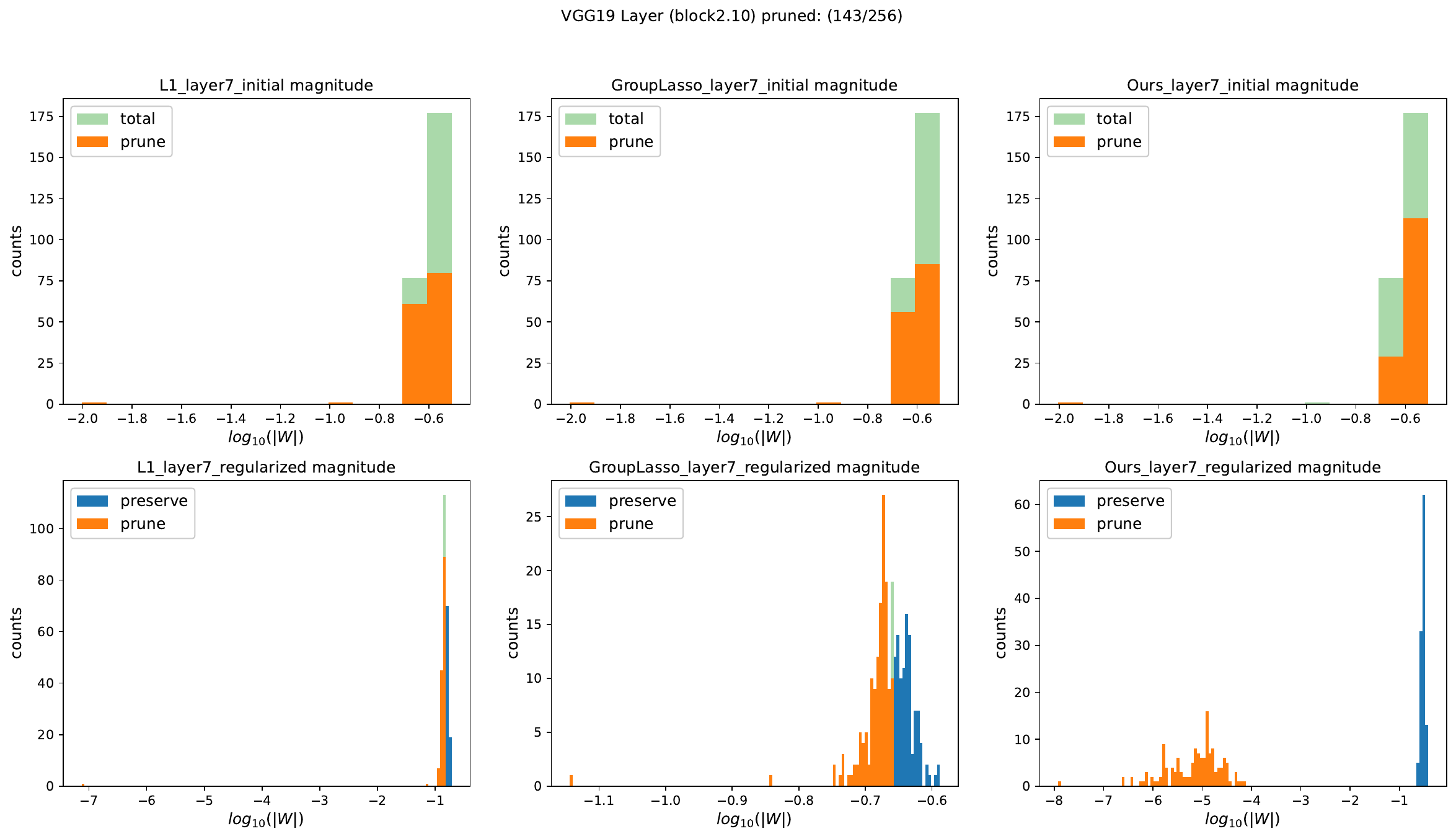}
    \caption{Initial magnitude and regularized magnitude of L1, Group Lass and our regularizer on 8th layer of VGG19 model. Each columns correspond to L1, Group Lasso, $\|DW\|_{2,1}$ respectively. First row is histogram of initial filter norms and second is historgram of regularized filter norms.}
\end{figure}
\begin{figure}[H]
    \centering
    \includegraphics[width=\textwidth]{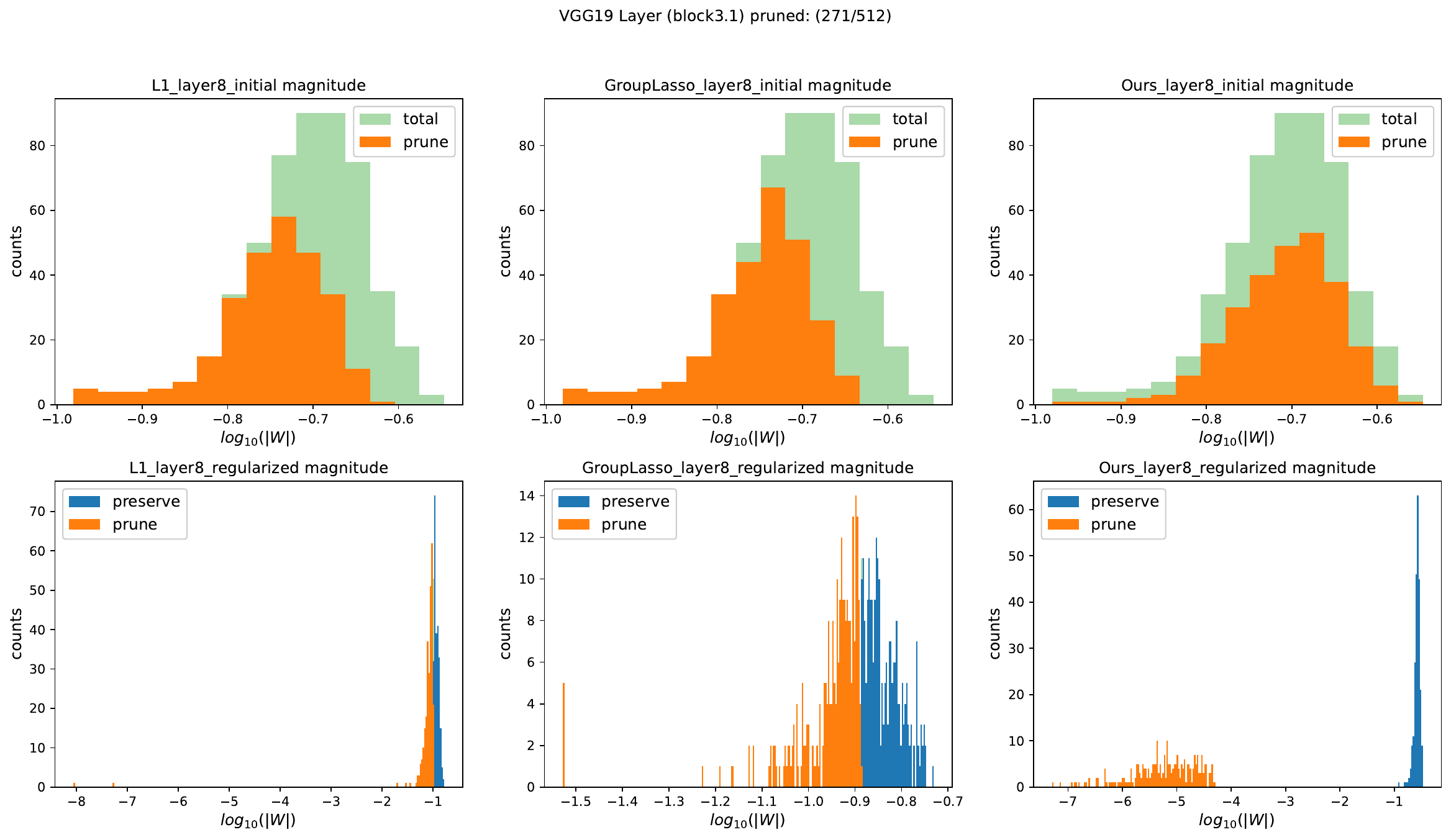}
    \caption{Initial magnitude and regularized magnitude of L1, Group Lass and our regularizer on 9th layer of VGG19 model. Each columns correspond to L1, Group Lasso, $\|DW\|_{2,1}$ respectively. First row is histogram of initial filter norms and second is historgram of regularized filter norms.}
\end{figure}\begin{figure}[H]
    \centering
    \includegraphics[width=\textwidth]{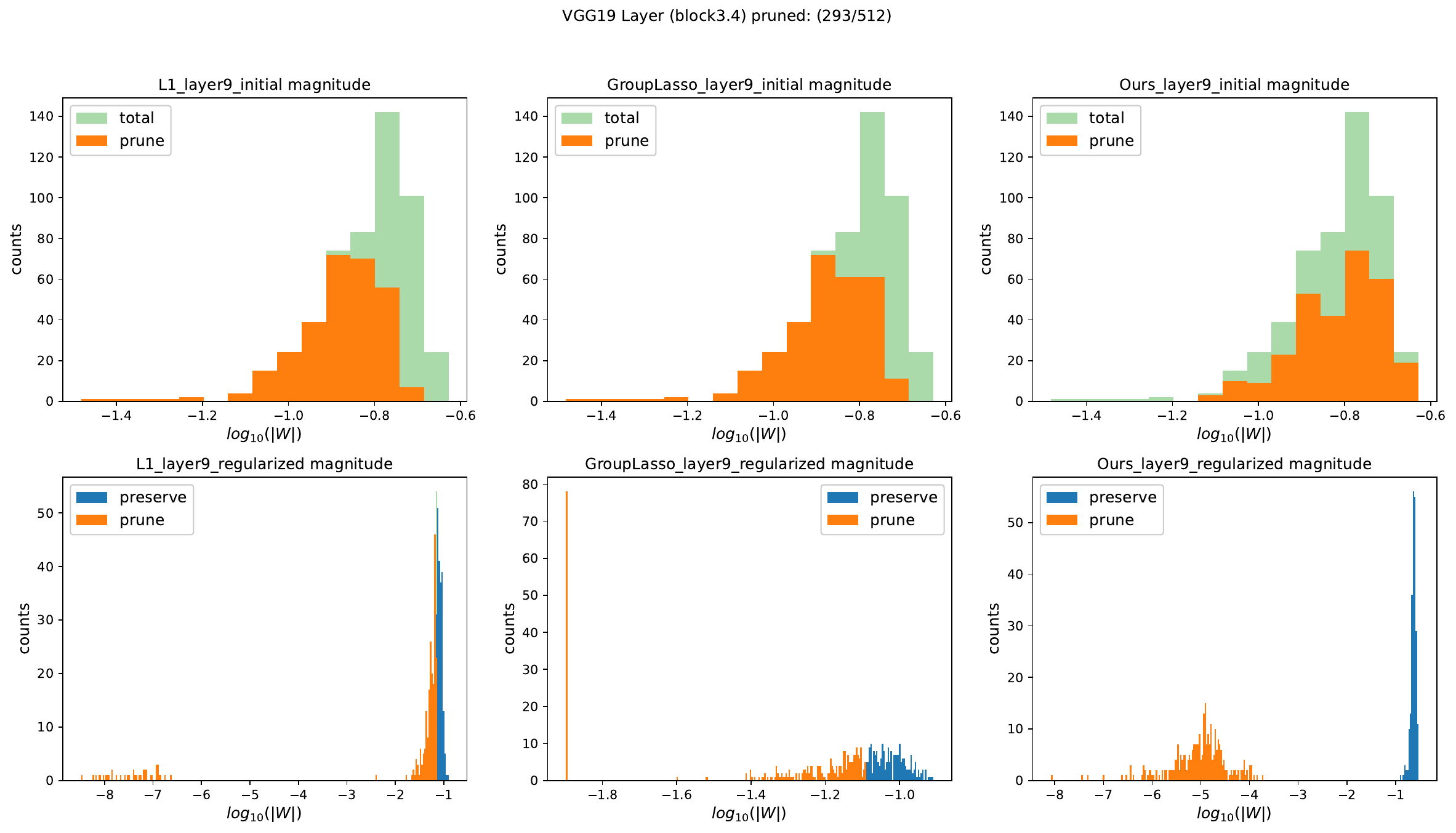}
    \caption{Initial magnitude and regularized magnitude of L1, Group Lass and our regularizer on 10th layer of VGG19 model. Each columns correspond to L1, Group Lasso, $\|DW\|_{2,1}$ respectively. First row is histogram of initial filter norms and second is historgram of regularized filter norms.}
\end{figure}

\begin{figure}[H]
    \centering
    \includegraphics[width=\textwidth]{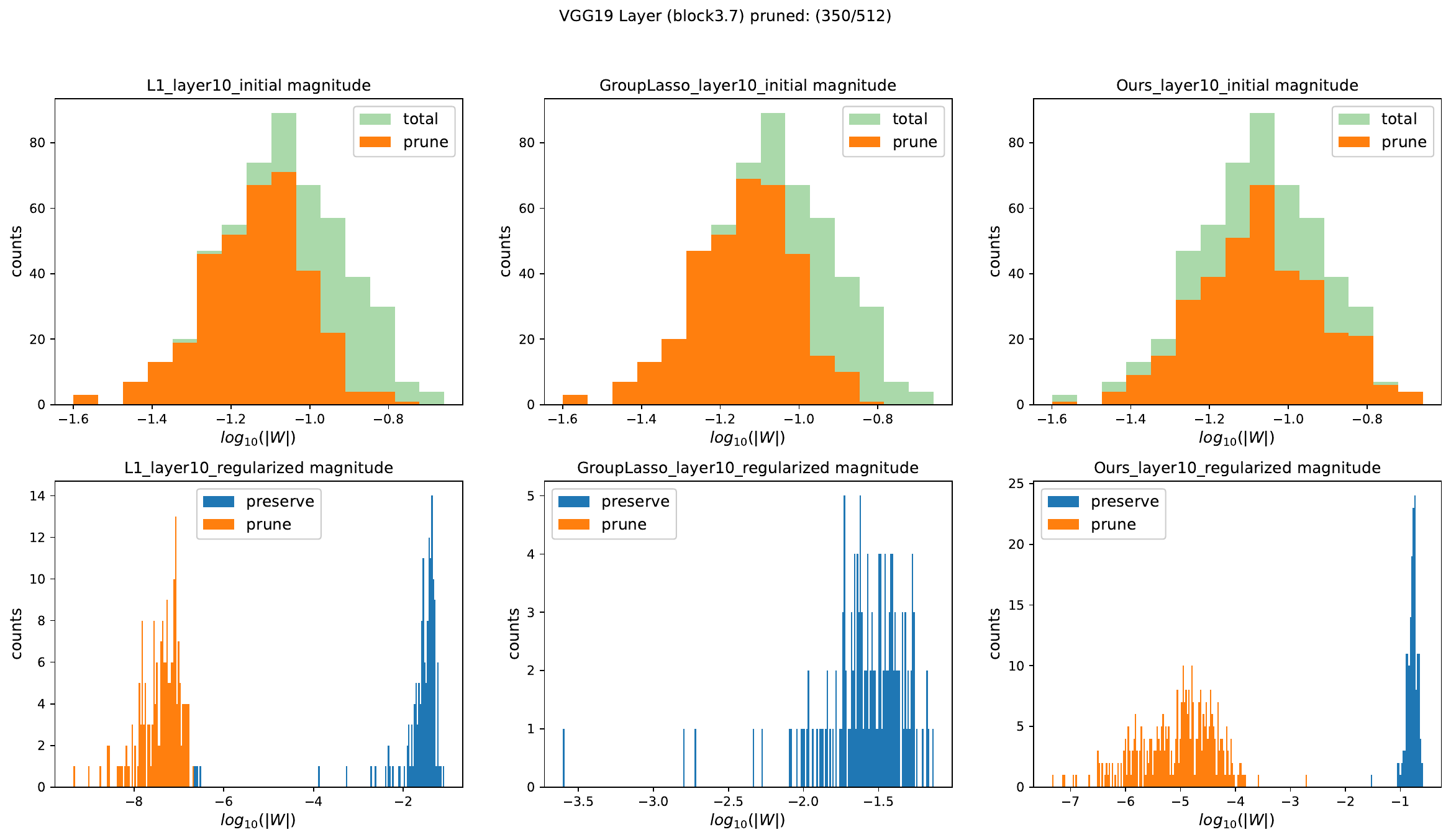}
    \caption{Initial magnitude and regularized magnitude of L1, Group Lass and our regularizer on 11th layer of VGG19 model. Each columns correspond to L1, Group Lasso, $\|DW\|_{2,1}$ respectively. First row is histogram of initial filter norms and second is historgram of regularized filter norms.}
\end{figure}
\begin{figure}[H]
    \centering
    \includegraphics[width=\textwidth]{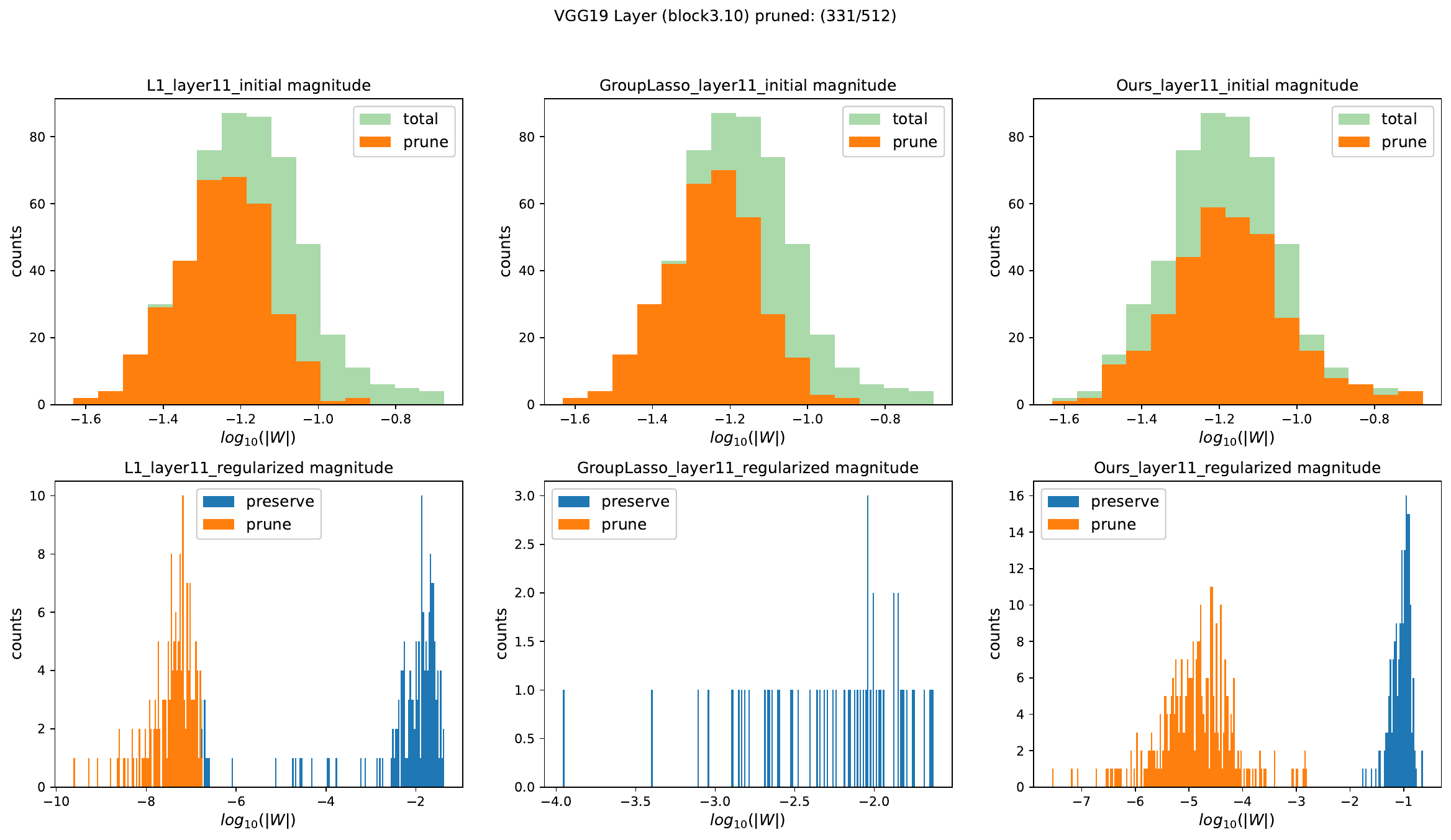}
    \caption{Initial magnitude and regularized magnitude of L1, Group Lass and our regularizer on 12th layer of VGG19 model. Each columns correspond to L1, Group Lasso, $\|DW\|_{2,1}$ respectively. First row is histogram of initial filter norms and second is historgram of regularized filter norms.}
\end{figure}
\begin{figure}[H]
    \centering
    \includegraphics[width=\textwidth]{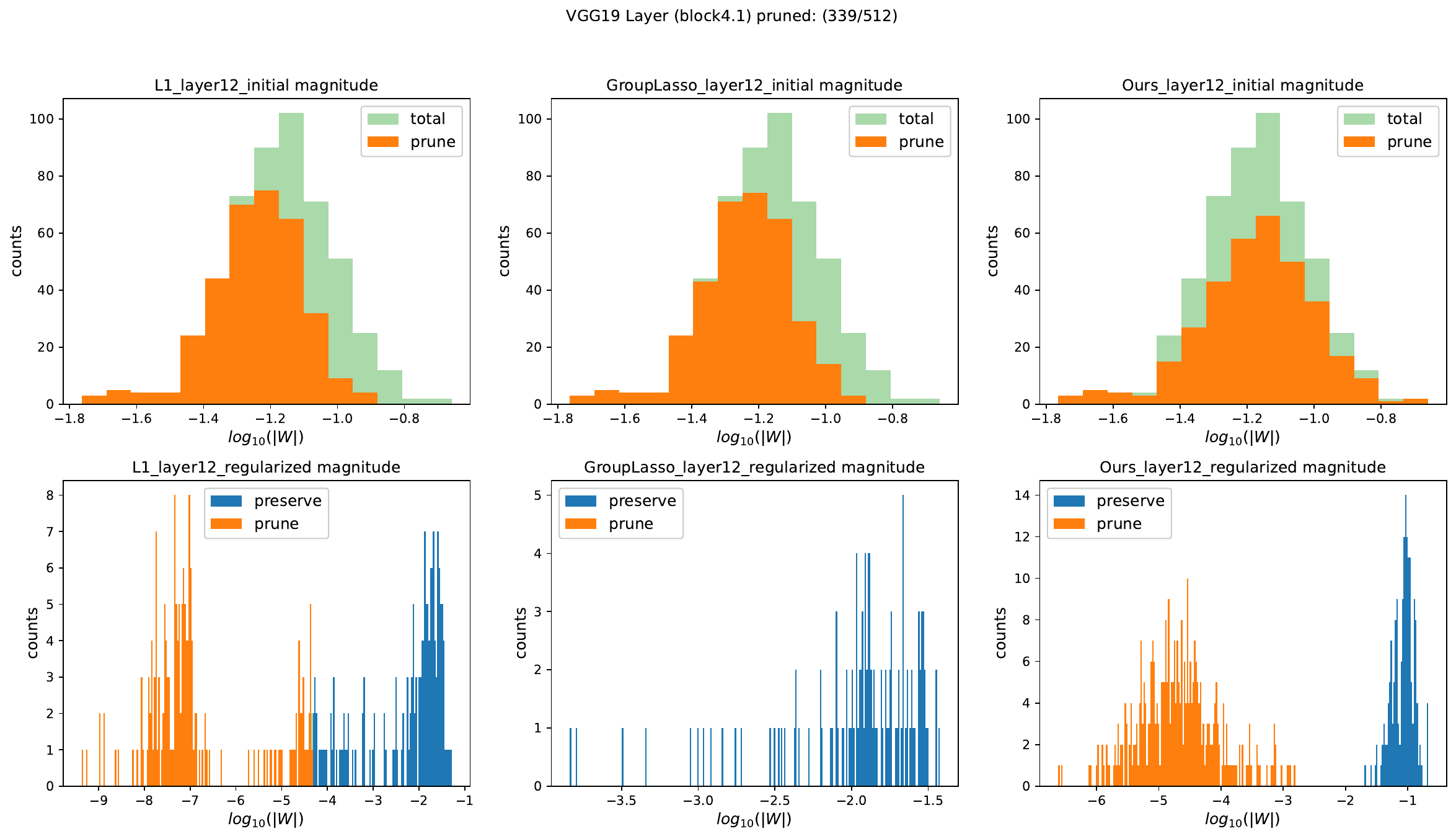}
    \caption{Initial magnitude and regularized magnitude of L1, Group Lass and our regularizer on 13rd layer of VGG19 model. Each columns correspond to L1, Group Lasso, $\|DW\|_{2,1}$ respectively. First row is histogram of initial filter norms and second is historgram of regularized filter norms.}
\end{figure}
\begin{figure}[H]
    \centering
    \includegraphics[width=\textwidth]{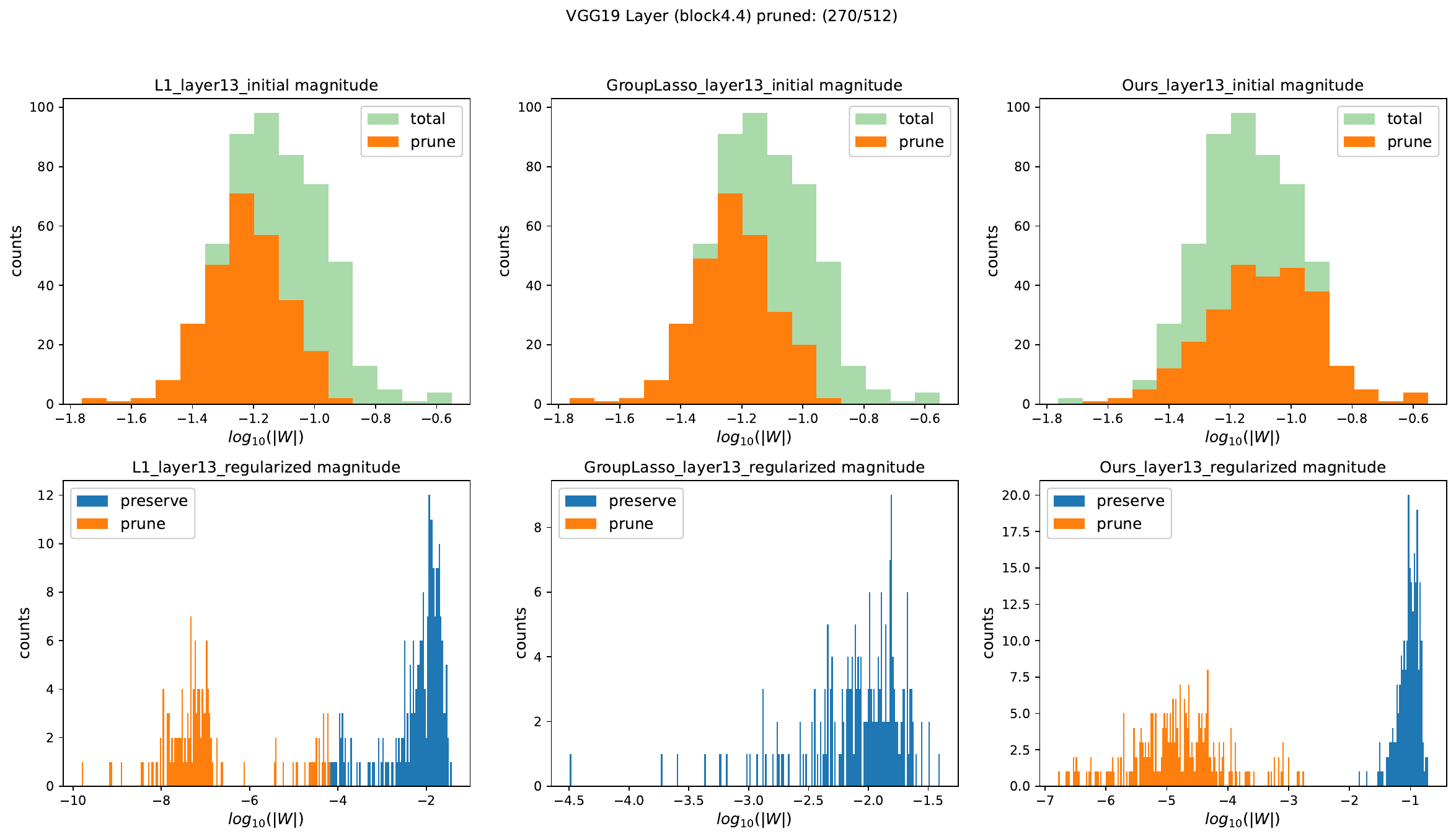}
    \caption{Initial magnitude and regularized magnitude of L1, Group Lass and our regularizer on 14th layer of VGG19 model. Each columns correspond to L1, Group Lasso, $\|DW\|_{2,1}$ respectively. First row is histogram of initial filter norms and second is historgram of regularized filter norms.}
\end{figure}\begin{figure}[H]
    \centering
    \includegraphics[width=\textwidth]{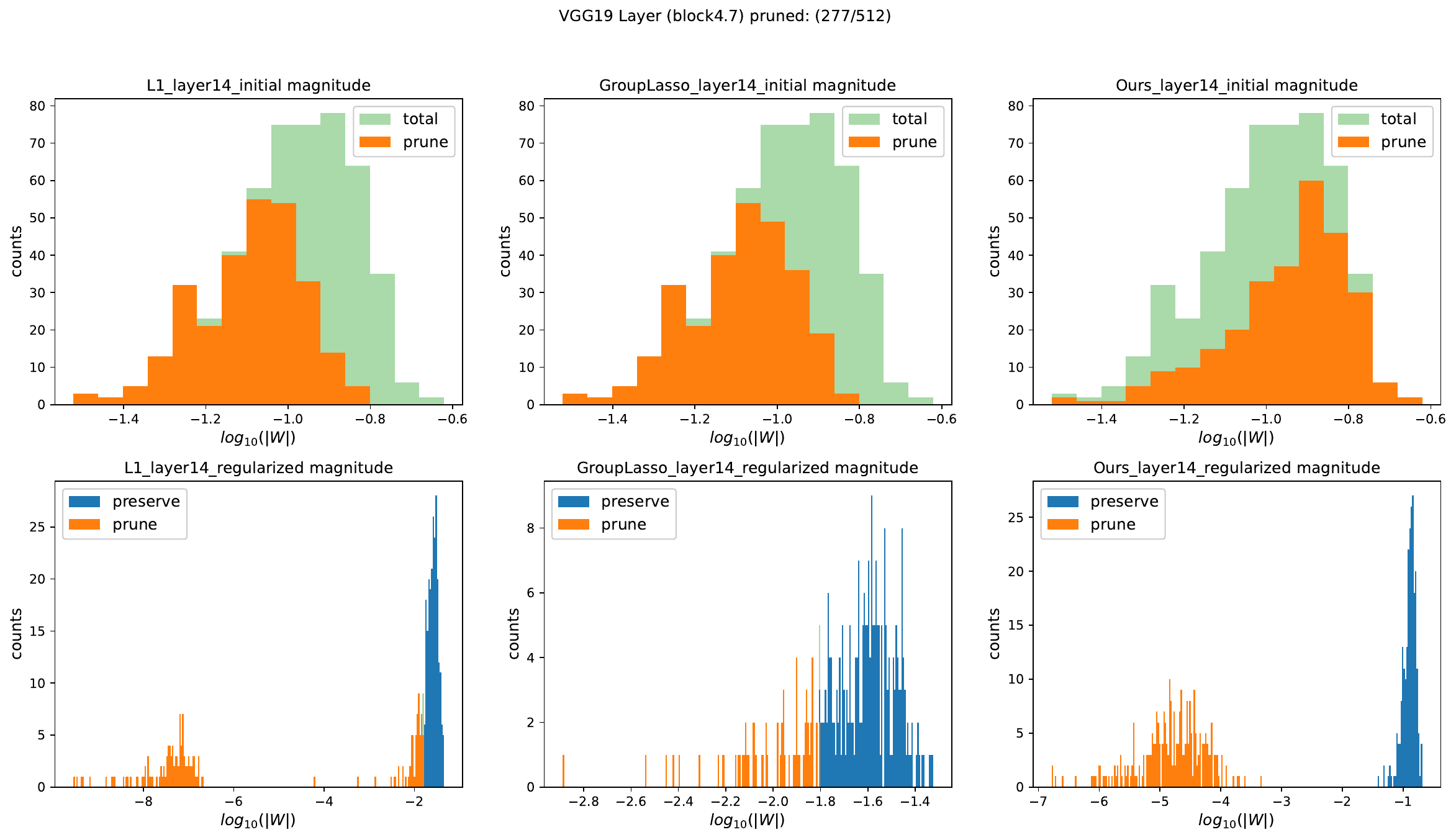}
    \caption{Initial magnitude and regularized magnitude of L1, Group Lass and our regularizer on 15th layer of VGG19 model. Each columns correspond to L1, Group Lasso, $\|DW\|_{2,1}$ respectively. First row is histogram of initial filter norms and second is historgram of regularized filter norms.}
\end{figure}

\begin{figure}[H]
    \centering
    \includegraphics[width=\textwidth]{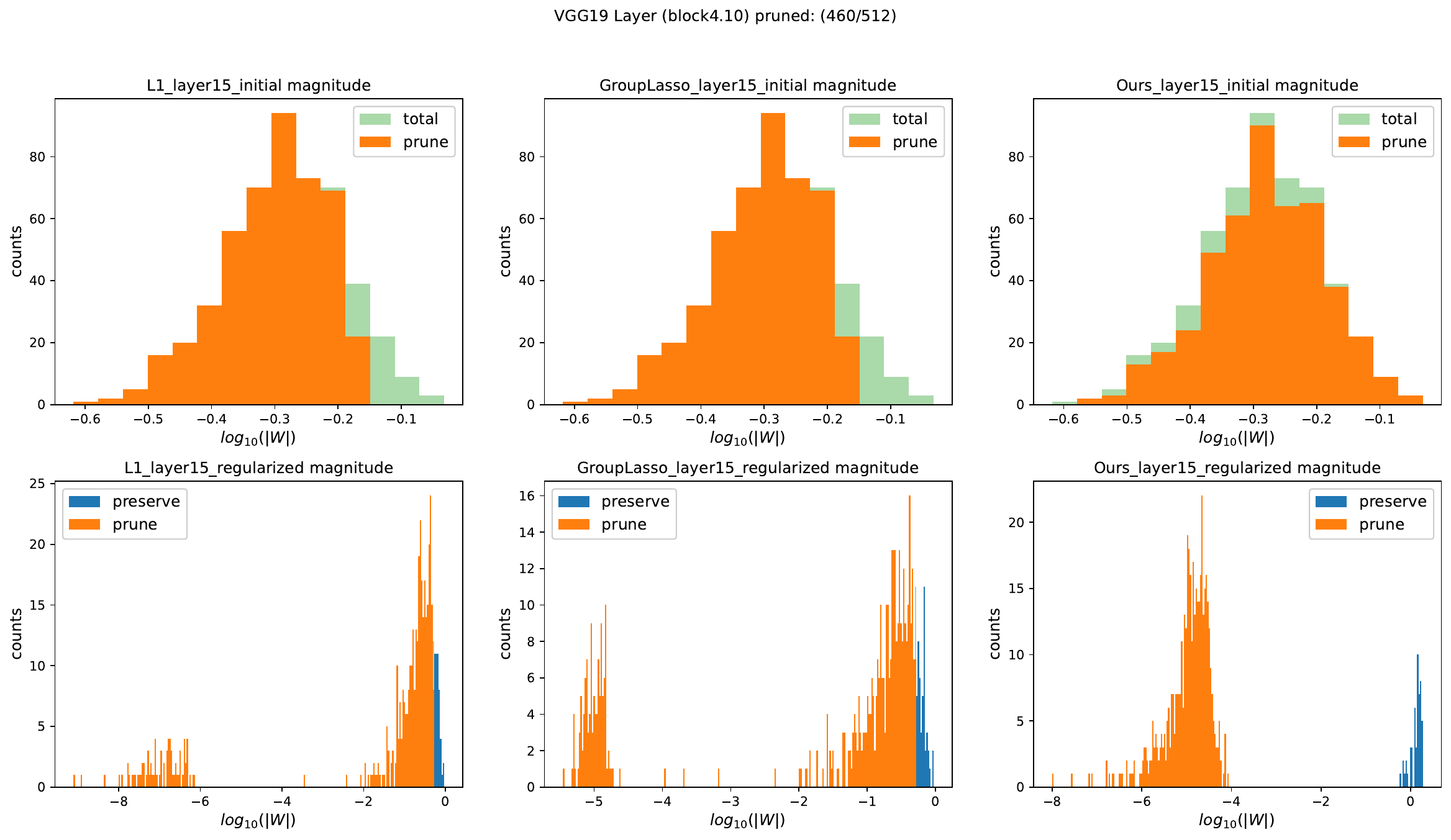}
    \caption{Initial magnitude and regularized magnitude of L1, Group Lass and our regularizer on 16th layer of VGG19 model. Each columns correspond to L1, Group Lasso, $\|DW\|_{2,1}$ respectively. First row is histogram of initial filter norms and second is historgram of regularized filter norms.}
\end{figure}
\newpage

\end{document}